\documentclass{article}
\usepackage{iclr2026_conference,times}
\usepackage{natbib}
\usepackage{xspace}
\usepackage{booktabs}
\usepackage{amsmath}
\usepackage{amssymb}
\usepackage{booktabs}
\usepackage{multirow}
\usepackage{enumitem}
\usepackage{wrapfig, tabularx}
\usepackage{caption}
\usepackage[T1]{fontenc}
\usepackage[table]{xcolor}
\usepackage[most]{tcolorbox}
\usepackage{url}
\usepackage{xurl}
\usepackage{algorithm}
\usepackage{algpseudocode}
\usepackage{capt-of}
\usepackage{fvextra}

\definecolor{promptframe}{HTML}{D5DDE8}
\definecolor{promptbg}{HTML}{F7F9FC}
\definecolor{prompttext}{HTML}{3A4450}
\definecolor{prompttitle}{HTML}{5B7290}
\fvset{breaklines=true, breaksymbolleft={}, fontsize=\scriptsize,
  baselinestretch=1.15, formatcom=\color{prompttext}}

\newtcolorbox{promptbox}[1][]{
  enhanced, breakable,
  colback=promptbg, colframe=promptframe,
  boxrule=0.5pt, arc=6pt,
  left=10pt, right=10pt, top=6pt, bottom=6pt,
  before skip=12pt, after skip=12pt,
  drop fuzzy shadow=black!6, #1}

\newcommand{\method}{\textsc{ConRAG}\xspace}
\newcommand{\spm}{\mathord{\vcenter{\hbox{$\scriptstyle\pm$}}}}

\newtcolorbox{casebox}[1]{%
  enhanced, breakable,
  colback=gray!4, colframe=black!70, colbacktitle=gray!15, coltitle=black,
  boxrule=0.6pt, arc=2pt,
  left=6pt, right=6pt, top=4pt, bottom=4pt,
  fonttitle=\bfseries\small, fontupper=\small,
  title={#1},
}

\title{\method: 
\\Lightweight inference of multi-hop relations}
\author{
  Anonymous Author(s) \\
  Affiliation \\
  \texttt{email}
}

\vspace{-0.8cm}
\author{Kilian Bänziger\thanks{Equal contribution. Correspondence to \texttt{kbaenziger@ethz.ch}.} \\
Agentic Systems Lab \\ ETH Zurich \\
\And Sonia Laguna\footnotemark[1] \\
ETH Zurich \\
\And Markus Kreft \\
Agentic Systems Lab \\ ETH Zurich \\
\And  Robert Jakob \\
Agentic Systems Lab \\ ETH Zurich  \\
\And Kevin O'Sullivan \\
Agentic Systems Lab \\ ETH Zurich  \\
 \And  Lasse B. Strand \\
Agentic Systems Lab, ETH Zurich \\ Norwegian University of Science and Technology \\
\And Julia E. Vogt \\
ETH Zurich  \\}

\iclrfinalcopy
\begin{document}
\maketitle
\fancyhead{}

\vspace{-0.6cm}
\begin{abstract} 
\vspace{-0.2cm}
Understanding how two entities are connected often requires tracing multi-hop relations across documents to identify intermediate entities and supporting evidence that explain a connection. This is a task that appears frequently in scientific research and other knowledge-intensive analyses. We formalise this setting as multi-hop relation inference: given two known endpoint entities, we aim to recover the bridge entities and evidence-grounded reasoning chains that connect them across a document corpus, and to generate an explanation grounded in the retrieved evidence. Existing multi-hop RAG systems typically seek an unknown answer entity rather than explicitly recovering the connection between two known endpoints and  graph-based approaches often rely on costly LLM-extracted knowledge graphs that limit scalability to large document collections. We introduce \method, which builds a lightweight entity–document graph from entity co-occurrence and LLM-based entity filtering. Its connective retrieval infers and semantically ranks paths between two endpoints. On MuSiQue and 2WikiMultiHopQA, \method consistently improves bridge entity and reasoning chain recovery over strong RAG baselines, while reducing graph-indexing token cost by up to roughly 1.5 orders of magnitude. Our results show that endpoint-constrained path retrieval provides an effective and index-efficient approach to evidence-grounded relation discovery. \looseness-1
\end{abstract}

\vspace*{-0.4cm}

\section{Introduction}

Retrieval Augmented Generation (RAG) \citep{lewis2020retrieval, gao2023retrieval} enables language models to access and reason over large document collections without requiring the entire corpus to fit within the input context. Rather than relying solely on knowledge encoded in model parameters, RAG systems retrieve a subset of relevant external information and provide it to a Large Language Model (LLM) as additional context for generation. By incorporating information not contained in the training corpus, RAG mitigates hallucinations by grounding generation in external knowledge~\citep{shi2023large, liu2024lost}. This improves answer quality and performance across knowledge-intensive natural language tasks~\citep{gao2023retrieval, fan2024survey}.\looseness-1

\vspace{-0.2cm}
\begin{wrapfigure}{l}{0.42\textwidth}
\vspace{-0.2cm}
  \vspace{-\baselineskip}
  \centering
  \includegraphics[width=\linewidth, trim={20pt 8pt 20pt 8pt}, clip]{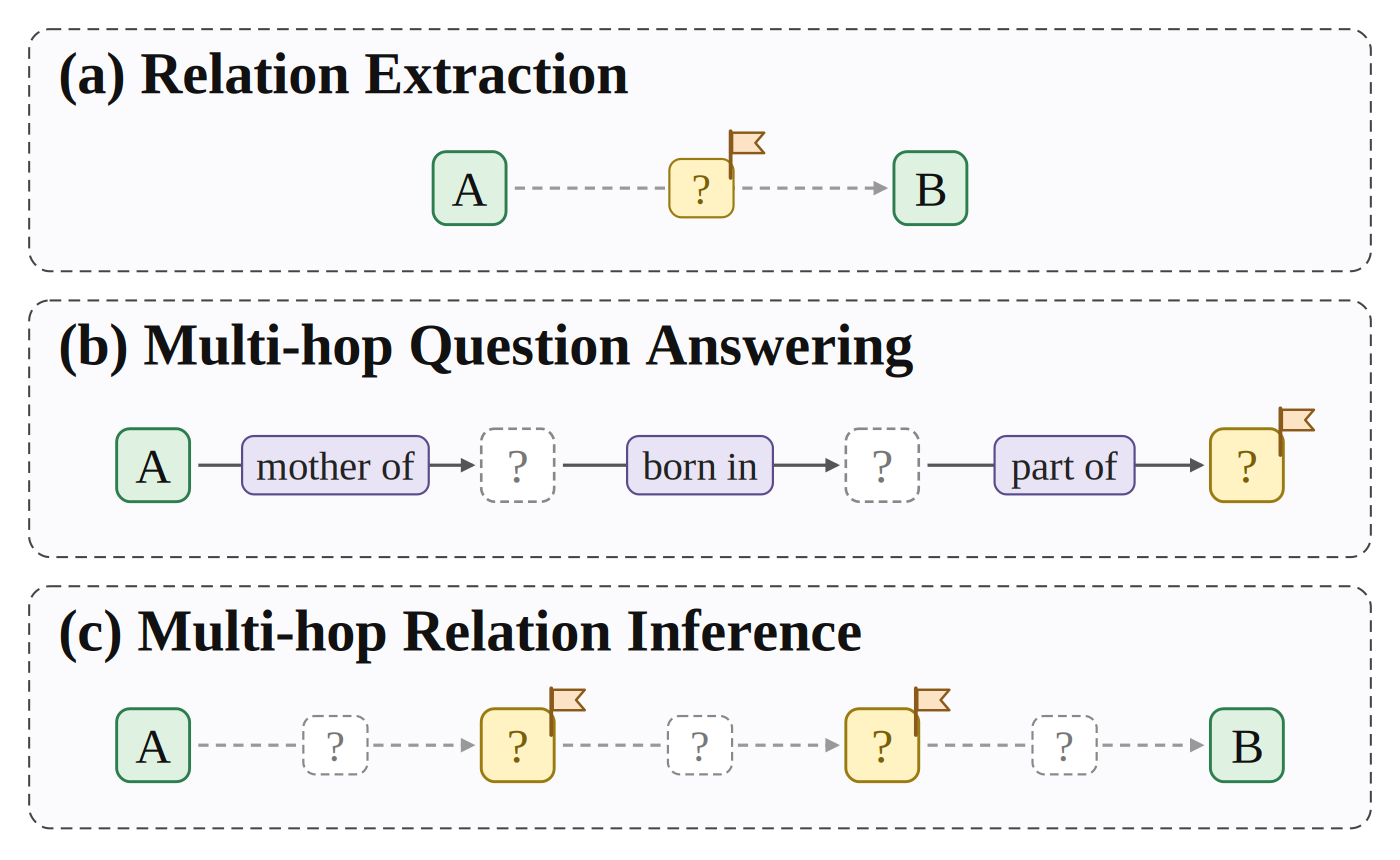}
  \vspace*{-0.7cm}
  \caption{Relation extraction, MHQA and multi-hop relation inference.\looseness-1}
  \vspace*{-0.4cm}
  \label{fig:differentiation}
\end{wrapfigure}

\paragraph{From simple relations to chain retrieval.}
Extracting entities and their relations from documents is important for real-world applications such as knowledge discovery~\citep{peng-etal-2017-cross}, contextualisation~\citep{edge2024local, guo2024lightrag}, scientific research and question answering~\citep{fader2014open}. These relations have been studied in different forms~\citep{niklaus2018survey,
han2020more}. Some are stated explicitly, such as ``Caesar was born in Rome''~\citep{zhang2017position}. Others are inferred by combining stated relations, for example when two parent links imply a grandparent relation~\citep{yao2019docred, yao2021codred, cabot2021rebel}. Prior work targets this challenge, including multi-hop relation extraction across documents \citep{wang2022entity, lu2023multi, son2023explore}. These fall under the task of relation extraction as shown in Figure~\ref{fig:differentiation} \textit{(a)}. \textit{(b)} and \textit{(c)} involve more complex connections that may only be expressed through a chain of intermediate entities. A researcher may be linked to a company through a co-author who founded a startup later acquired by that company. Multi-hop question answering (MHQA)~\citep{mavi2024multi} (Figure~\ref{fig:differentiation}\textit{(b)})  involves answering questions like \textit{"Where was the author of Marcinkus born?"} where these relations and an anchor entity \textit{(Marcinkus)} are given in the question~\citep{trivedi2022musique, ho2020constructing, yang2018hotpotqa}. We instead treat the chain itself and not a single relation label as the target: Given two known entities, we retrieve the bridge entities and links connecting them (Figure~\ref{fig:differentiation} \textit{(c)}). We call this \emph{multi-hop relation inference}.\looseness-1

\vspace{-0.2cm}
\paragraph{Why existing multi-hop RAG does not solve it.} 
As outlined, MHQA requires reasoning across multiple steps~\citep{mavi2024multi}. State-of-the-art RAG methods for MHQA either interleave retrieval with stepwise LLM reasoning \citep{trivedi2023interleaving, sun2024think} or retrieve over an LLM-extracted knowledge graph (KG) \citep{gutierrez2024hipporag, gutierrez2025rag}. Despite this, they are not explicitly designed to reconstruct the chains connecting two given entities. Their retrieval objectives are typically relevance-based or answer-oriented rather than endpoint-constrained: Documents may be scored individually or graph neighborhoods expanded toward an unknown answer, without requiring the evidence to form a connected path between two known endpoints. Consequently, essential bridge documents with weak query similarity may be missed. \looseness-1 

\vspace{-0.2cm}
\paragraph{Can multi-hop relation inference be natively supported in RAG?} 
To close the gap and enable RAG based multi-hop relation inference, we introduce Connective Retrieval Augmented Generation (\method)
, a graph-based RAG framework designed to recover multi-hop connections between two known entities. \method constructs an entity-document graph from a corpus of documents and retrieves paths between endpoint-associated documents, surfacing the intermediate entities and evidence needed to explain their connection. To evaluate this setting, we give the standard MHQA benchmarks MuSiQue~\citep{trivedi2022musique} and 2WikiMultiHopQA~\citep{ho2020constructing} a new interpretation. Question and answer entities become the known endpoint pairs, while annotated bridge entities and decomposition steps become the target chains. \method outperforms RAG baselines in bridge entity recovery
and improves reasoning chain recall and precision, while reducing indexing token cost by up to 1.5 orders of magnitude. More precisely, our contributions are the following:\looseness-1


\begin{figure}[t]
  \centering
  \vspace*{-0.4cm}
  \includegraphics[width=0.85\textwidth]{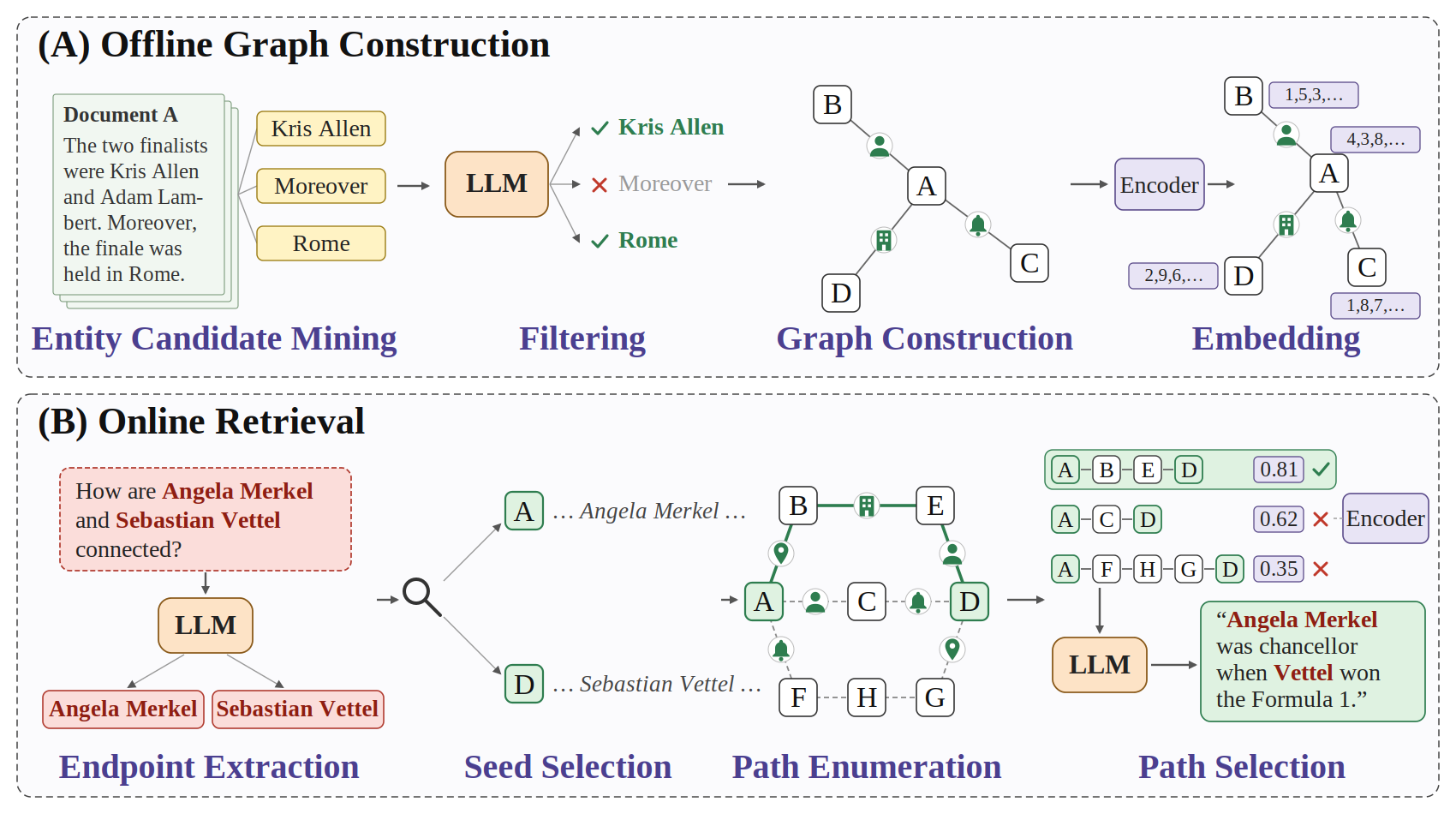}
  \vspace*{-0.3cm}
\caption{\textbf{Overview of \method} (outlined in Section~\ref{sec:method}).
\textbf{(A) Offline graph construction:} An LLM filters entity candidates mined from the documents, the remaining entities link documents in a graph and a dense encoder embeds each document. \textbf{(B) Online retrieval:} Anchor and target are extracted from the question. Seed documents are retrieved for both. Paths are enumerated and scored by question-document similarity. The top path documents form the context to explain the relation.\looseness-1}
  \label{fig:conrag_method}
  \vspace*{-0.5cm}
\end{figure}

\vspace*{-0.1cm}
\begin{itemize}[nosep, leftmargin=*]
\item 
We formalise multi-hop relation inference and introduce new benchmarks derived from MuSiQue and 2WikiMultiHopQA, with known endpoint pairs, bridge entities, and reasoning chains as targets. This way, we provide a testbed and recipe for future research on relation discovery.\looseness-1

\item We introduce \method, the first graph-based RAG framework designed to recover connections between known entities, with a token efficient indexing solution via LLM entity filtering.

\item We show that \method outperforms all evaluated RAG baselines on the studied benchmarks.
\end{itemize}

\section{Related Work}
\vspace{-0.1cm}
\paragraph{Explaining relations.}
Explaining how two entities are related, rather than merely detecting that they are, has a long history across data management, the Semantic Web and NLP~\citep{anyanwu2003rho,faloutsos2004fast,swanson1986fish}, motivated by exploratory search and by users' needs for supporting evidence in high-stakes domains~\citep{bhatia2018tell,pirro2015explaining}. Works like REX~\citep{fang2011rex} infer and rank explanatory paths for relevance over pre-existing KGs, while~\citet{voskarides2015learning} combine KG information with sentences retrieved from a document corpus to explain a relation. In contrast, our method operates on raw text, induces its own entity-document graph, and uses an LLM to explain the recovered relation chains in prose.

\vspace*{-0.2cm}
\paragraph{Relation extraction.} 
A broader line of work infers direct relations between entities within a sentence or document~\citep{zhao2024comprehensive}. DocRED~\citep{yao2019docred} extends this to document-level extraction, and models such as REBEL~\citep{cabot2021rebel} and UIE~\citep{lu2022unified} generate relation triplets end-to-end. Moreover, CodRED~\citep{yao2021codred} moves to the cross-document setting, where the relation between an entity pair must be inferred from evidence spread over several documents. Graph-based methods such as MR.COD~\citep{lu2023multi}, a multi-hop evidence retriever for cross-document relation extraction, and PILOT~\citep{son2023explore}, construct entity-document paths and train dense retrievers to select evidence for a downstream classifier that predicts a predefined relation label between endpoints. Path inference over KGs follows a similar formulation. From path-ranking random walks~\citep{lao2011random} to reinforcement-learning walkers~\citep{xiong2017deeppath}, multi-hop paths over curated triples serve as evidence for predicting a direct relation between two given entities. Importantly, these methods are not framed as RAG systems: Retrieved paths serve as evidence for a relation extractor or classifier, rather than as context for an LLM to generate an evidence-grounded explanation. Therefore, in these settings the output is a direct relation label, and intermediate entities and documents are auxiliary. Our setting inverts this: The label is dropped, and the chain of bridge entities and evidence becomes the target. \method introduces the relation inference setting within RAG and builds on the evidence-path retrieval formulation of \citet{lu2023multi}, while removing its reliance on gold mention annotations and a task-specific trained retriever in favour of an efficient, LLM-filtered entity vocabulary.

\vspace*{-0.275cm}
\paragraph{Retrieval Augmented Generation.}
RAG~\citep{lewis2020retrieval} conditions generation on documents retrieved from an external corpus. It was proposed for knowledge-intensive tasks and later adopted to reduce unsupported claims by LLMs and to inject knowledge absent from pretraining~\citep{gao2023retrieval, fan2024survey}. Standard pipelines retrieve documents by semantic similarity.
Graph-based RAG addresses this by organising the corpus as a graph. GraphRAG~\citep{edge2024local} and LightRAG~\citep{guo2024lightrag} extract entities and relations with an LLM and retrieve over the resulting KG. PathRAG~\citep{chen2026pathrag} alters this by pruning paths for retrieval. MiniRAG~\citep{fan2025minirag} builds a lightweight graph of chunks and extracted entities. LinearRAG~\citep{zhuang2026linearrag} bypasses relation extraction altogether, connecting SpaCy-extracted entities through shared sentences and documents and retrieving by propagating activation from the entities mentioned in the query. More cost-efficient RAGs are outlined in Appendix~\ref{app:ext_rel_work}. \method is closest to MiniRAG and LinearRAG in using relation-free entity-document graph and to PathRAGs retrieval. It differs in the following respects. First, the entity vocabulary is inferred from corpus co-occurrence statistics and LLM filtering. Second, compared to PathRAG, retrieval is not pruning based and traverses a graph without relations. We search for paths between two known endpoints in a construction-efficient graph and rank path candidates on average document similarity with an encoder.\looseness-1

\vspace*{-0.275cm}
\paragraph{Multi-hop reasoning.}
MHQA benchmarks such as HotpotQA, 2WikiMultiHopQA, and MuSiQue~\citep{yang2018hotpotqa, ho2020constructing, trivedi2022musique} require composing facts across documents. Some MHQA questions contain an anchor entity and a sequence of relations leading to an answer through unknown bridge entities. Existing methods that address this challenge fall into several families. Iterative retrieval interleaves retrieval with LLM reasoning as in IRCoT~\citep{trivedi2023interleaving}, or lets an LLM traverse a KG~\citep{sun2024think}. The resulting chain from the question's anchor to its answer is a byproduct. Chain-aware retrievers such as MDR~\citep{xiong2020answering} and CogQA~\citep{ding2019cognitive} learn to retrieve sequences of documents or to expand an entity graph from the question, to reach an unknown answer. Finally, graph-based methods like  HippoRAG and HippoRAG 2~\citep{gutierrez2024hipporag, gutierrez2025rag} leverage graph algorithms to surface important relational information.
None of these families explicitly targets the setting of two known entities with an unlabeled, evidence-grounded chain between them. \method makes the chain of bridge entities and evidence the output itself, and assumes no predefined relation vocabulary.\looseness-1

\section{\method: Connective Retrieval Augmented Generation}

\label{sec:method}

We propose \method, a RAG framework for multi-hop relation inference that improves inference and understanding of indirect relations between two entities.

\subsection{Problem formulation}
\label{sec:method:problem}

Let $\mathcal{D} = \{d_1,\dots,d_N\}$ be a corpus of $N$ documents. A question $q$ names an \emph{anchor} entity $a$ and a \emph{target} entity $b$ like \textit{Rome} and \textit{Pythagoras}. The system
must produce output $\mathcal{O}(q)$\looseness-1
\begin{equation}
  \mathcal{O}(q) = (\mathcal{C}, y),
  \qquad
  \mathcal{C} = \bigl( u_1 \!\to\! u_2,\; \dots,\; u_{k-1} \!\to\! u_{k} \bigr),
  \label{eq:output}
\end{equation}
where $\mathcal{C}$ is an ordered sequence of $k$ entities with $u_1 = a$ and $u_{k} = b$, written as
directed links.  The links outline the indirect relation between $a$ and $b$. $y$ is a prose explanation of the indirect relation. The interior entities are the \emph{bridges}. Both $\mathcal{C}$ and $y$ must be
entailed by a retrieved document set $\mathcal{R}(q) \subseteq \mathcal{D}$. Emitting $\mathcal{C}$ separately from $y$ makes the claimed reasoning path directly confirmable rather than inferable from free text.

\subsection{Entity qualification}
\label{sec:method:entities}
To construct the bipartite entity-document graph for retrieval, we first derive an entity vocabulary from document texts which then serves as a connector between documents. \method proceeds in the two steps below.

\vspace{-0.2cm}
\paragraph{Entity candidate mining.}
Two families of candidates are extracted from each document. The first is the set of lowercased and possessive stripped
unigrams. The second is the set of phrases of at most $n_{\max}$ tokens drawn from maximal sequences of
capitalised tokens. For each candidate $e$ we record the document set
$\mathcal{D}_e$ containing it and its document frequency $\mathrm{df}(e) = |\mathcal{D}_e|$ over the whole corpus. Candidates with $\mathrm{df}(e) < 2$ are discarded since a mention confined to one document cannot connect two documents.

\vspace{-0.2cm}
\paragraph{LLM-based entity filtering.}
Mined candidates are filtered by an LLM with instructions given in Appendix~\ref{app:llm_prompts_filtering} containing the capitalisation rate of the candidates across the corpus. It labels genuine named entities and discards other candidates in containment-aware batches. Candidates linked by contiguous token containment (\emph{united}
$\subset$ \emph{united states} $\subset$ \emph{united states of america}) are placed in the same batch of size $B$. Competing granularities of one name are thus adjudicated in a single context rather than in independent calls.

\subsection{Offline indexing}
\label{sec:method:index}

Building upon the entity vocabulary we construct an index used for the retrieval of relevant information during inference as shown in Figure~\ref{fig:conrag_method} \textit{(A)}. The index is a triple $\mathcal{I} = (\mathcal{D}, G, \mathbf{V})$: The corpus $\mathcal{D}$, a
bipartite entity-document graph $G$, and a document embedding matrix $\mathbf{V} \in
\mathbb{R}^{N \times r}$, where $r$ is the embedding dimension and whose row $\mathbf{v}_d$ holds the vector of document $d$. Before indexing, $\mathcal{D}$ is normalised by deduplicating pairs of titles and texts and collapsing whitespaces from documents in the datasets. 

This relation-free construction avoids committing to a predefined relation schema and eliminates the need to extract relation triples from every document. Entities act only as connectors between documents, while the original documents remain the source of evidence presented to the generator.

\vspace{-0.2cm}
\paragraph{Graph construction.}
To construct the graph, entities are first assigned to documents from the incidence recorded during mining (Sec.~\ref{sec:method:entities}). Moreover, if one entity is a contiguous token subsequence of another within a document, only the longer is kept. In this way a document is linked by the most specific name available to it. Based on this, we construct an undirected bipartite graph $G = (\mathcal{V}_{\mathcal{D}} \cup
\mathcal{V}_{\mathcal{E}}, E)$ where $\mathcal{V}_{\mathcal{D}}$ is the set of document nodes and $\mathcal{V}_{\mathcal{E}}$ is the set of entity nodes. An undirected edge exists if and only if an entity is assigned to a document. Together, all edges form the set of edges $E$. There are no document-document or entity-entity edges. 

\textbf{Document embedding.}
Each document is embedded from its title and text with a dense encoder to give
$\mathbf{V}$. These vectors serve two roles downstream, dense seeding and query-document support.

\textbf{Node canonicalisation.}
Entity nodes are canonicalised across the corpus. For this purpose entity strings from the vocabulary are embedded as well. Candidate pairs for canonicalisation are restricted to entities that share a word or whose initials are acronyms of the other. Nodes among these pairs whose cosine similarity exceeds a threshold $\theta$ are merged by the union-find algorithm~\citep{galler1964improved, tarjan1975efficiency}, and each group is represented by its member with the highest document frequency. Since entity embeddings are not used for retrieval they do not become part of $\mathbf{V}$.

\subsection{Online retrieval}
\label{sec:method:retrieval}

\setlength{\intextsep}{-13pt}

\begin{wrapfloat}{algorithm}[10]{r}{0.46\textwidth}
\caption{\method Online Retrieval}
\label{alg:conrag_retrieval}

\footnotesize
\begin{algorithmic}[1]
\Require question $q$, graph $G$, embeddings $\mathbf{V}$, corpus $\mathcal{D}$
\Ensure retrieved context $\mathcal{R}(q)$

\State Extract anchor $a$ and target $b$ from $q$
\State Retrieve dense seeds $\mathcal{R}_{\mathrm{seed}}(q)$
\State Ensure coverage of both endpoints
\State Enumerate paths between endpoint-side seeds
\State Score paths by question-document similarity
\If{no valid path is found}
    \State Apply shortest-path search
\EndIf
\State Select top paths as $\mathcal{R}_{\mathrm{path}}(q)$
\State \Return $\mathcal{R}_{\mathrm{seed}}(q)
       \cup \mathcal{R}_{\mathrm{path}}(q)$
\end{algorithmic}

\vspace{-0.2\baselineskip}
\end{wrapfloat}

During retrieval \method aims to collect information that is meaningful to
reconstruct indirect connections between entities.The retrieval procedure is summarised in Algorithm~\ref{alg:conrag_retrieval}
and illustrated in Figure~\ref{fig:conrag_method} \textit{(B)}.

\paragraph{Endpoint extraction.}
Entity mentions are extracted from question $q$ by an LLM. The first
two mentions are taken as the anchor $a$ and the target $b$. If fewer than two mentions are found, $q$ itself is used as the seeding signal and the endpoint constraint is skipped.

\paragraph{Seeding.}
Seed documents are retrieved densely. Documents are ranked by $\cos(\mathbf{v}_q, \mathbf{v}_d)$,
where $\mathbf{v}_q$ is the embedding of the whole question and  $\mathbf{v}_d$ is a document embedding. The top $k_{\mathrm{seed}}$ documents form $\mathcal{R}_{\mathrm{seed}}(q)$. 
A path from $a$ to $b$ requires seed documents covering both endpoints. Top-$k$ dense retrieval can return one-sided seed sets, thus we restrict $\mathcal{R}_{\mathrm{seed}}(q)$ to documents containing $a$ or $b$ and inject a matching document from the corpus when an endpoint is missing from the semantic search. If this fails, we fall back to a search with the entity embedding against the documents. Finally, if this does not yield documents containing an entity, we use the top-$k_{\mathrm{fallback}}$ documents of the question embedding search for the missing endpoint.

\paragraph{Path enumeration.}
Following~\citet{lu2023multi}, we enumerate paths by depth-first search over $d \to e \to d'$ transitions from the
anchor-side seeds. A path of $L$ documents is
\begin{equation}
  P = \bigl( (d_1,\dots,d_L),\, (e_1,\dots,e_{L-1}) \bigr),
  \qquad e_i \text{ the entity shared by } d_i \text{ and } d_{i+1},
\end{equation}
with all $d_i$ distinct and $L \leq L_{\max}$, where $L_{\max}$ is the
maximum path length in documents. Entities whose degree $\deg(e) = |\mathcal{D}_e|$ exceeds a $p_{\mathrm{hub}}$ quantile of the degree distribution are removed from the traversal adjacency. Enumeration is constrained. A path is retained
only if node $d_L$ lies in the target-side seed pool.

\setlength{\intextsep}{12pt plus 2pt minus 2pt}   
\begin{wrapfigure}[4]{r}{0.36\textwidth}
\vspace{-0.8cm}
\begin{equation}
S(P) =
\frac{1}{L}\sum_{i=1}^{L}
\cos(\mathbf{v}_q,\mathbf{v}_{d_i}).
\label{eq:pathscore}
\end{equation}
\end{wrapfigure}
\paragraph{Path selection.}
We score paths by the mean query support of documents as in Eq.~\ref{eq:pathscore} \citep{lu2023multi}. Scored paths are bucketed by length. The $k_{\mathrm{path}}$ highest-scoring paths are retained for each length in a target set $\mathcal{L}$, and their documents form $\mathcal{R}_{\mathrm{path}}(q)$. If no enumerated path has a length in $\mathcal{L}$, retrieval falls back to a single shortest path. The candidate with the highest score under Eq.~\ref{eq:pathscore} is emitted alone. The retrieval procedure is bounded by the maximum path length $\mathcal{L}_{\max}$, while hub suppression limits expansion through high-degree entities. Together, these constraints keep connective search tractable while the shortest path fallback preserves connectivity when the pruned graph is too sparse.

\paragraph{Context assembly.}
The generator LLM receives $\mathcal{R}(q) = \mathcal{R}_{\mathrm{seed}}(q) \cup
\mathcal{R}_{\mathrm{path}}(q)$. Seeds supply endpoint evidence and remain available when the graph is
locally disconnected. Paths supply the intermediate documents. Which documents carry the relations is left to the generator.
\vspace{-0.2cm}
\subsection{Constructing Multi-hop Relation Inference Benchmarks}
\label{sec:benchmark_construction}

Multi-hop question answering datasets with question decompositions $S$ can be
repurposed for multi-hop relation inference. Given an MHQA question-answer pair with a
linear reasoning chain, we identify the entity from which the decomposition
starts as the \emph{anchor} $a$ and use the original answer as the
\emph{target} $b$. If the anchor is not explicitly given in the
decomposition, we extract from its first decomposition question with an LLM instructed by a prompt given in Appendix~\ref{app:llm_prompts_extraction}.

\setlength{\intextsep}{+1pt}

\begin{wrapfloat}{algorithm}[12]{r}{0.53\textwidth}
\caption{MHQA to Multi-hop Relation Inference}
\label{alg:relation_dataset}

\footnotesize
\begin{algorithmic}[1]
\Require MHQA decomposition $S$, answer $b$, supporting documents $\mathcal{D}_{\mathrm{gold}}$
\Ensure relation-inference example $(q,a,b,Z,c^\star,\mathcal{D}_{\mathrm{gold}})$

\State Identify anchor $a$ from the first step of $S$
\State Collect intermediate answers $Z=[z_1,\ldots,z_{h-1}]$
\State Set answer $b$ as the target entity
\State Build chain
       $c^\star = (a \to z_1 \to\cdots\to z_{h-1} \to b)$
\State Form relation question $q$ from endpoints $(a,b)$
\State \Return $(q,a,b,Z,c^\star,\mathcal{D}_{\mathrm{gold}})$
\end{algorithmic}
\end{wrapfloat}

For an $h$-hop decomposition where $h$ is the number of reasoning hops, the intermediate answers define the ordered bridge
entities $Z=[z_1,\ldots,z_{h-1}]$. Together with the anchor and target, they
form the gold relation chain consisting of intermediate links.
\[
c^\star = (a \rightarrow z_1 \rightarrow \cdots
     \rightarrow z_{h-1} \rightarrow b).
\]
We then replace the original question with a relation question $q$ that names
only the two endpoints, e.g., ``How are $a$ and $b$ related?''. The resulting
example consists of the endpoint pair $(a,b)$, the bridge entities $Z$, 
the gold chain $c^\star$, and the original supporting documents
$\mathcal{D}_{\mathrm{gold}}$.
We restrict the construction to linear chains with a single anchor, since
branching questions require recovering multiple interacting relation chains.
Algorithm~\ref{alg:relation_dataset} summarises the general
procedure.\looseness-1

\setlength{\intextsep}{12pt plus 2pt minus 2pt}
\vspace{-0.2cm}
\section{Experimental Setup}

\paragraph{Datasets.}
We instantiate the benchmark construction procedure from Section~\ref{sec:benchmark_construction} on the validation sets of MuSiQue~\citep{trivedi2022musique} and 2WikiMultiHopQA~\citep{ho2020constructing}. For MuSiQue, we retain the linear \textit{2hop}, \textit{3hop1}, and \textit{4hop1} compositions. 
This yields 2,066 examples which comprise 1,252 two-hop, 568 three-hop, and 246 four-hop questions. For 2WikiMultiHopQA, we retain the \textit{compositional} and \textit{inference} categories, yielding 6,785 two-hop examples. Retrieval operates over pooled, deduplicated validation corpora rather than per-question context bundles, comprising approximately 26.3k documents for MuSiQue and 56.7k for 2WikiMultiHopQA. Dataset-specific processing and anchor extraction details are provided in Appendix~\ref{app:exp_setup}.\looseness-1

\vspace{-0.2cm}
\paragraph{Evaluation.} 
We introduce three derived metrics for multi-hop relations benchmarking, \textit{Bridge contains match accuracy (BCM)}, \textit{chain recall (CR)} and \textit{chain precision (CP)}. 
BCM measures the share of gold bridge entities included in a response.
CR and CP measure the share of gold links recovered, as well as the share of generated links that are gold.
We additionally evaluate response quality against the baselines.
We follow LightRAG \citep{guo2024lightrag} and use their prompt to instruct an LLM judge comparing \method's responses against baselines in terms of \textit{comprehensiveness}, \textit{diversity}, \textit{empowerment} and \textit{overall} response quality. 
Finally, we report \method's average win rate against each baseline. 
For quantitative results, we evaluate three 1,000-question samples drawn with different seeds. While limiting the question corpus to reduce generation cost, we still use the full document corpus of the validation set for retrieval to show indexing efficiency.\looseness-1

\vspace{-0.2cm}
\paragraph{Baselines} 
We  compare \method against four RAG baselines. We include \textit{Na\"{i}ve RAG} \citep{lewis2020retrieval} as a semantic search baseline. Moreover, \method is compared against \textit{LightRAG} as being one of the most capable graph models regarding qualitative metrics \citep{guo2024lightrag}. Third, we include \textit{PathRAG} since it uses a path-based retrieval similar to \method~\citep{chen2026pathrag}. Finally, \textit{HippoRAG 2} is included as a strong multi-hop reasoning baseline~\citep{gutierrez2025rag}. With this setup, we compare \method against methods that leverage relation-rich KGs, which can provide fine-grained information about the underlying corpus for generation. Baseline implementation details are outlined in Appendix~\ref{app:baselines_imp}.

\vspace{-0.2cm}
\paragraph{Implementation Details.}
All methods use OpenAI's \textit{text-embedding-3-small} model for semantic embedding. HippoRAG 2, LightRAG, and PathRAG use \textit{GPT-4o-mini} for graph indexing.
\method uses \textit{GPT-4o} for entity filtering and generation. The values of \method's parameters are listed in Appendix~\ref{app:exp_setup} Table~\ref{tab:params} and kept fixed across datasets. All methods use the same generation prompt (Appendix~\ref{app:llm_prompts_generation}), with method-specific retrieved context. As the number of retrieved documents from \method averages to 10.8 we limit the retrieval of na\"{i}ve RAG, HippoRAG and LightRAG to the top 10 documents as well. PathRAG retrieval varies and is not restrictable and therefore remains as is. We use \textit{GPT-4o-mini} as LLM judge for qualitative criteria.

\vspace*{-0.2cm}
\section{Results}
\vspace*{-0.1cm}
Our evaluation is organised around four questions:
\textit{(i)} Does \method infer multi-hop relations more accurately than traditional and graph-based RAG baselines?
\textit{(ii)} Does this translate into higher-quality responses as judged by a qualitative evaluation?
\textit{(iii)} How much of the gain stems from retrieval alone, independent of LLM-based generation?
\textit{(iv)} How does \method's efficiency compare in terms of indexing and inference cost to the baselines?

\subsection{Relation Inference}
\begin{table}[h!]
\centering
\caption{Bridge Contains Match Accuracy (BCM), Chain Recall (CR), and Chain Precision (CP) in \%
on MuSiQue and 2WikiMultiHopQA (mean\,$\spm$\,std over 3 seeds).
Best per column in \textbf{bold}, second-best \underline{underlined}.}
\label{tab:bridge-chain}
\setlength{\tabcolsep}{4pt}
\scriptsize
\renewcommand{\arraystretch}{0.82}
\resizebox{\linewidth}{!}{%
\begin{tabular}{lcccccc}
\toprule
& \multicolumn{3}{c}{\textit{MuSiQue}} & \multicolumn{3}{c}{\textit{2WikiMultiHopQA}} \\
\cmidrule(lr){2-4}\cmidrule(lr){5-7}
Method
& \textit{BCM} & \textit{CR} & \textit{CP}
& \textit{BCM} & \textit{CR} & \textit{CP} \\
\midrule
Na\"{i}veRAG
& $59.2{\spm}0.6$ & $19.6{\spm}0.4$ & $27.4{\spm}1.1$
& $61.6{\spm}1.5$ & $33.4{\spm}2.4$ & $\underline{40.1{\spm}2.3}$ \\
LightRAG
& $58.0{\spm}1.7$ & $20.2{\spm}0.3$ & $28.3{\spm}0.1$
& $\underline{72.1{\spm}0.2}$ & $\underline{40.2{\spm}1.0}$ & $35.7{\spm}1.1$ \\
PathRAG
& $58.3{\spm}0.8$ & $17.2{\spm}0.3$ & $33.6{\spm}1.3$
& $64.9{\spm}0.2$ & $30.4{\spm}1.5$ & $35.8{\spm}1.8$ \\
HippoRAG\,2
& $\underline{59.6{\spm}2.2}$ & $\underline{23.3{\spm}0.6}$ & $\mathbf{36.4{\spm}1.7}$
& $67.0{\spm}1.0$ & $\underline{40.2{\spm}0.9}$ & $34.9{\spm}2.1$ \\
\method
& $\mathbf{71.2{\spm}0.7}$ & $\mathbf{29.5{\spm}0.5}$ & $\underline{34.3{\spm}1.1}$
& $\mathbf{81.0{\spm}1.4}$ & $\mathbf{49.4{\spm}1.3}$ & $\mathbf{48.1{\spm}0.9}$ \\
\bottomrule
\end{tabular}
}
\end{table}

In Table~\ref{tab:bridge-chain} we present results regarding performance on relation inference. It is worth to highlight that MuSiQue poses a more challenging task for relation inference since it contains relations with up to 3 intermediate steps, while 2WikiMultiHopQA only contains relations with one intermediate step for inference. The results allow to make the following observations:

\vspace{-0.2cm}
\paragraph{Superior performance on multi-hop relation inference.} Due to its mechanism which is designed to connect entities, \method achieves results that underpin its ability to surface relations. Most prominently, this is visible through BCM which highlights a methods awareness regarding the existence of bridge entities. \method outperforms baselines across both datasets. Moreover, \method proves strong ability to infer reasoning chain links. Deltas over $6.2$ points on MuSiQue and $9.2$ points on 2WikiMultiHopQA in chain recall to HippoRAG 2 indicate that even against systems laid out for precise multi-hop reasoning, \method is capable to infer and understand relations more effectively.\looseness-1

\begin{wrapfigure}[16]{r}{0.5\textwidth}
\vspace{-0.9\baselineskip}
\centering
\includegraphics[width=\linewidth]{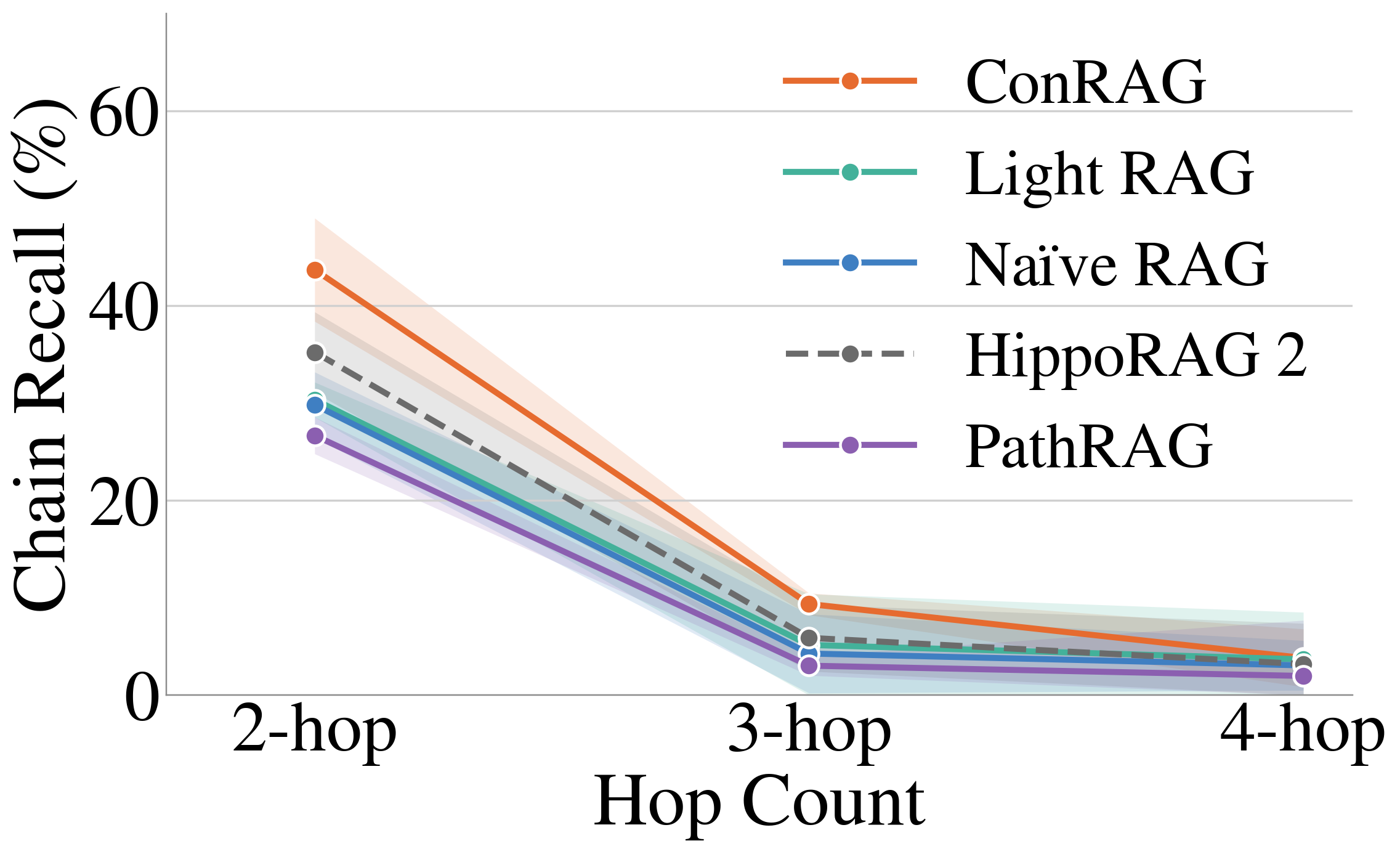}
\vspace{-0.6cm}
\caption{\textbf{Decreasing chain recall (\%).} All models show lower chain recall as the number of reasoning hops increases. Bands indicate two-sided 99\% t-confidence intervals.}
\label{fig:decrease_with_hopcount}
\vspace{-0.5\baselineskip}
\end{wrapfigure}

\paragraph{Longer relation chains remain challenging.} As shown in Figure~\ref{fig:decrease_with_hopcount}, we can derive that with rising hop count, inference becomes more difficult. 
Performance decreases substantially with hop count across all methods. Nevertheless, \method retains the highest bridge-recovery at every hop count (Appendix~\ref{app:add_results} Table~\ref{tab:hop_breakdown_br_cr_cp}), outperforming the strongest baseline . Overall, this can be seen as a result of an increase in possible chain connections with a rise in hopcount due to more combinatorial options for path assembly.
When controlling for hop count, \method consistently performs better than baselines across both benchmarks, indicating performance robustness. While all models show decreasing performance over hop count, we can also observe that baseline performance varies strongly between the two-hop questions of the different datasets. As outlined in Appendix~\ref{app:add_results} Table~\ref{tab:hop_breakdown_br_cr_cp}, deviations for baseline models rise up to $10.6$ points for Na\"{i}ve RAG on BCM and up to $9.9$ points for chain recall (LightRAG).

\vspace{-0.2cm}
\paragraph{Ranking of baselines varies.}
Notably, our experiments show that baseline models do not keep a consistent ranking across metrics and datasets among each other. Only HippoRAG 2 shows a tendency in being the strongest baseline in chain recall, which indicates that its capabilities for multi-hop reasoning are partly transferable to multi-hop relation inference. For BCM and chain precision, there are no consistent top performers among the baseline models. In contrast, \method consistently performs above baseline level.  

\subsection{Qualitative Performance}

\begin{table}[h!]
  \centering
  \scriptsize
\caption{\textbf{Pairwise qualitative win rate (\%) of \method against each baseline.} (reported as mean $\pm$ 95\% Wilson half-width) 
For every question and criterion, the LLM judge selects either \method or the
baseline as the winner. Therefore, the two win rates sum to 100\%, and only
\method's share is reported. Values above 50\% indicate that \method wins the
majority of comparisons and are shown in \textbf{bold}. Compr. = comprehensiveness,
Divers. = diversity, and Empow. = empowerment.}
  \label{tab:winrates}

  \setlength{\tabcolsep}{3.5pt}
  \resizebox{\linewidth}{!}{%
  \begin{tabular}{
@{}l
ccc>{\columncolor{gray!10}}c
@{\hspace{8pt}}
ccc>{\columncolor{gray!10}}c
@{}
}
\toprule
\multirow[c]{2}{*}[-0.6ex]{\method{} vs. Baseline}
& \multicolumn{4}{c}{\textit{MuSiQue}}
& \multicolumn{4}{c}{\textit{2WikiMultiHopQA}} \\
\cmidrule(lr){2-5} \cmidrule(lr){6-9}
& Compr. & Divers. & Empow. & Overall
& Compr. & Divers. & Empow. & Overall \\
\midrule
Na\"{i}ve RAG
  & \textbf{54.3 $\spm$ 3.1} & \textbf{50.1 $\spm$ 3.1} & \textbf{53.5 $\spm$ 3.1} & \textbf{53.5 $\spm$ 3.1}
  & \textbf{57.7 $\spm$ 3.1} & 49.2 $\spm$ 3.1 & \textbf{57.2 $\spm$ 3.1} & \textbf{57.2 $\spm$ 3.1} \\
LightRAG
  & \textbf{69.6 $\spm$ 2.9} & \textbf{65.2 $\spm$ 3.0} & \textbf{69.6 $\spm$ 2.9} & \textbf{69.6 $\spm$ 2.9}
  & \textbf{60.1 $\spm$ 3.0} & \textbf{55.4 $\spm$ 3.1} & \textbf{59.8 $\spm$ 3.1} & \textbf{59.8 $\spm$ 3.1} \\
PathRAG
  & \textbf{63.1 $\spm$ 3.0} & \textbf{60.3 $\spm$ 3.0} & \textbf{62.0 $\spm$ 3.0} & \textbf{62.0 $\spm$ 3.0}
  & \textbf{62.1 $\spm$ 3.0} & \textbf{55.0 $\spm$ 3.1} & \textbf{61.7 $\spm$ 3.0} & \textbf{61.8 $\spm$ 3.0} \\
HippoRAG\,2
  & \textbf{65.8 $\spm$ 3.0} & \textbf{64.2 $\spm$ 3.0} & \textbf{65.5 $\spm$ 3.0} & \textbf{65.7 $\spm$ 3.0}
  & \textbf{61.7 $\spm$ 3.0} & \textbf{52.1 $\spm$ 3.1} & \textbf{61.0 $\spm$ 3.0} & \textbf{61.1 $\spm$ 3.0} \\
\bottomrule
\end{tabular}
  }
\end{table}

We measure qualitative performance of \method by reporting win rates on relation questions against baselines over one question sample for each dataset. In Table~\ref{tab:winrates} we note win rates across both datasets and all four baselines. \method attains the higher overall win rate. Na\"{i}ve RAG generates the second most valuable answers in terms of our metrics and even outperforms \method on diversity for 2WikiMultiHopQA. Counterintuitively, the win margin widens against graph-based RAGs. Most clearly, \method wins more than twice as much against LightRAG over the MuSiQue dataset. This suggests that the improvement is not merely an effect of retrieving more but also less distracting context. Where LightRAG and PathRAG provide comprehensive generation prompts (Table~\ref{tab:prompt_comparison_sizes}) important details can get lost. \method's retrieval works differently. Answers are drawn from a tighter, higher-precision evidence set, which trades breadth of stated content for relevance.

\subsection{Indexing and Inference Efficiency}

\begin{wraptable}[11]{r}{0.67\textwidth}
\vspace{-1.0\baselineskip}
\centering
\small
\setlength{\tabcolsep}{2.5pt}
\caption{Average prompt tokens ($\spm$ std) \& index construction tokens.
\textbf{Bold} indicates lowest cost among graph-based RAGs.\looseness-1}
\label{tab:prompt_comparison_sizes}
\vspace{-0.3cm}
\renewcommand{\arraystretch}{0.95}
\begin{tabular}{@{}lcccc@{}}
\toprule
\multirow{2}{*}{Model}
& \multicolumn{2}{c}{\textit{MuSiQue}}
& \multicolumn{2}{c}{\textit{2WikiMultiHopQA}} \\
\cmidrule(lr){2-3} \cmidrule(lr){4-5}
& Prompt & Construction & Prompt & Construction \\
\midrule
PathRAG
& $10.312\spm3.547$ & 263.25\,M & $11.546\spm3.494$ & 564.37\,M \\
LightRAG
& $7.484\spm2.786$ & 263.25\,M & $8.427\spm3.031$ & 564.37\,M \\
Na\"{i}ve RAG
& {\boldmath$2.242\spm344$} & -- & {\boldmath$2.542\spm699$} & -- \\
HippoRAG\,2
& $2.245{\spm}349$ & 29.71\,M & $2.854\spm833$ & 65.27\,M \\
\method
& $2.261{\spm}983$ & \textbf{3.90\,M} & $3.301\spm1.707$ & \textbf{9.63\,M} \\
\bottomrule
\end{tabular}
\vspace{-0.2\baselineskip}
\end{wraptable}

To design a RAG which is useful in a setting with large document corpora and
regular requests, efficiency in terms of token cost for construction and
generation has to be considered. \method is designed to reduce token
consumption during indexing to a minimum. Our results in Table~\ref{tab:prompt_comparison_sizes} show that \method reduces indexing cost in comparison to
LightRAG and PathRAG on MuSiQue and 2WikiMultiHopQA by roughly 1.5 orders of magnitude. Furthermore, \method also clearly undercuts the comparatively lightweight HippoRAG\,2.\looseness-1

Moreover, \method saves tokens during generation. Context heavy RAGs like LightRAG and PathRAG consume more tokens by design to provide valuable context for answers regarding qualitative metrics. \method instead provides more compact context for the single purpose of surfacing and understanding multi-hop relations. It thereby attains a comparable generation token footprint as HippoRAG\,2 which is designed for compact context retrieval to provide correct MHQA responses.
Figure~\ref{fig:entity_funnel} in Appendix~\ref{app:ablations} illustrates the source of
\method's indexing efficiency. The document-frequency and LLM filtering
stages remove most candidate entities while largely preserving the bridge
entities required for path retrieval. This produces a substantially smaller
entity vocabulary before graph construction.

\newpage
\vspace{-0.3cm}
\subsection{Retrieval}

\vspace{-0.2cm}
\begin{wrapfigure}[12]{r}{0.5\textwidth}
  \vspace{-13pt}
  \centering
  \small
  \setlength{\tabcolsep}{1pt}
  \renewcommand{\arraystretch}{1}
  \captionof{table}{Document Recall (\%) on benchmarks
  (mean\,$\spm$\,std over 3 seeds). Best in \textbf{bold},
  second-best \underline{underlined}.}
  \label{tab:doc-recall}
  \vspace*{-0.3cm}
  \begin{tabularx}{\linewidth}{l >{\centering\arraybackslash}X >{\centering\arraybackslash}X}
    \toprule
    Model & \textit{MuSiQue} & \textit{2WikiMultiHopQA} \\
    \midrule
    Na\"{i}ve RAG & $61.1{\spm}0.6$             & $67.0{\spm}0.3$ \\
    LightRAG      & $60.7{\spm}0.9$             & $69.0{\spm}0.1$ \\
    PathRAG       & $62.9{\spm}0.8$ & $72.2{\spm}0.8$ \\
    HippoRAG\,2   & $\underline{64.2{\spm}2.3}$             & $\underline{79.0{\spm}0.3}$ \\
    \method       & $\mathbf{70.2{\spm}0.5}$    & $\mathbf{84.1{\spm}0.6}$ \\
    \bottomrule
  \end{tabularx}
\end{wrapfigure}

Finally, we isolate retrieval from generation to showcase improvement independent of the generator: Table~\ref{tab:doc-recall} reports
document recall, i.e.\ how much gold evidence reaches the generator before any
answer is produced. Since \textsc{ConRAG} retrieves 10.8 documents on average,
we report R@10 for baselines with a fixed retrieval size (all except PathRAG).
\textsc{ConRAG} achieves the highest recall on both benchmarks.
Because the document budget is comparable, the improvement does not result from
retrieving more. Moreover, retrieval
\parfillskip=0pt\par\parfillskip=0pt plus 1fil\relax
\vspace{-\parskip}
\begin{wrapfigure}[22]{r}{0.5\textwidth}
  \vspace{\dimexpr-\intextsep-\baselineskip\relax}    
  \raggedleft
  \includegraphics[width=\linewidth]{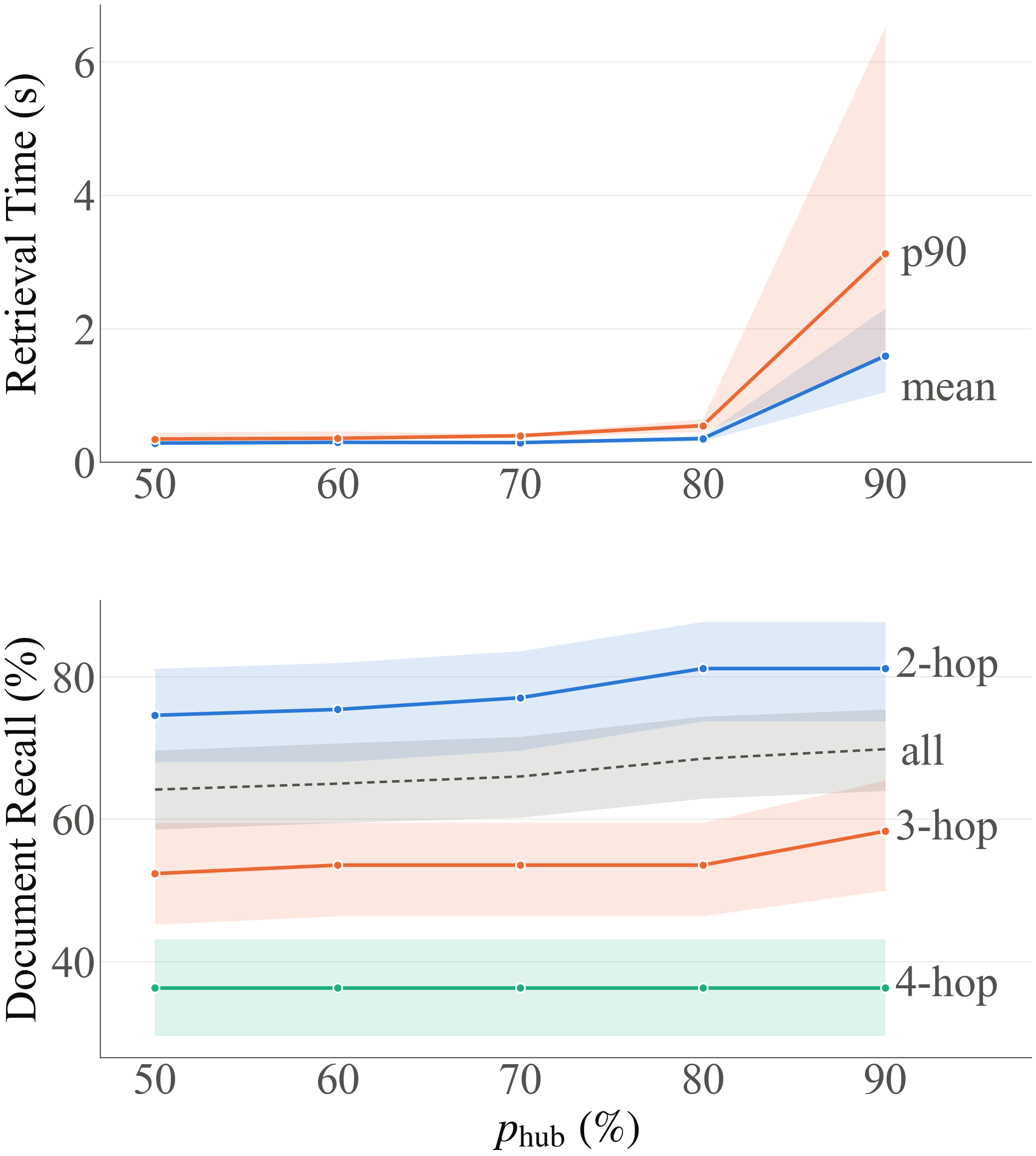}
  \vspace{-0.5cm}
  \caption{\textbf{Hub suppression.}
  $p_{\mathrm{hub}}$ and retrieval time in seconds and document recall in \%. Shaded bands are 95\% bootstrap confidence intervals.}
  \label{fig:hub_supression}
\end{wrapfigure}
\noindent
depends on hub suppression. Raising $p_{\text{hub}}$ steadily lifts recall while also increasing retrieval time due to increasing combinatorial complexity of path enumeration. As shown in Figure~\ref{fig:hub_supression} retrieval time grows strongly after passing $p_{\text{hub}}=0.8$ while not significantly improving overall document recall which results in a tradeoff between recall and retrieval time for \method. \looseness-1

\vspace{-0.3cm}
\section{Conclusion}

\vspace{-0.2cm}
In this paper, we introduced \method, a lightweight RAG framework for multi-hop relation inference between two known entities. \method constructs a relation-free entity-document graph using corpus co-occurrence and LLM-based entity filtering, and retrieves paths that explicitly connect the two endpoints through intermediate bridge entities and supporting evidence. We also formalised the task and derived evaluation benchmarks from MuSiQue and 2WikiMultiHopQA by treating annotated reasoning chains as targets between known endpoints. Across both benchmarks, \method improves bridge-entity and reasoning-chain recovery over strong RAG baselines, while also achieving higher document recall and substantially lower graph-indexing token requirements. These results
support endpoint-constrained path retrieval as an effective and index-efficient approach for evidence-grounded relation discovery. More broadly, this work motivates greater attention to multi-hop relation inference as a first-class retrieval problem. Recovering \emph{how} known entities are connected is important for scientific discovery, literature analysis, and other knowledge-intensive settings where the connecting evidence matters as much as the endpoint answer. Our results suggest that this capability does not require richly extracted knowledge graphs: A lightweight, relation-free representation combined with endpoint-constrained retrieval already provides a strong and efficient foundation for relation discovery.

\vspace{-0.3cm}
\paragraph{Limitations \& Future Work.}
Our dataset inherits the gold decompositions of MuSiQue and 2WikiMultiHopQA as the only reference chains. Two entities may, however, be connected through the corpus by several valid routes. Currently, every deviation from the reference is scored as an error, although alternative relations can be valid. Reported CR, CP, and BCM should therefore be read as conservative lower bounds. A second limitation is retrieval time: The cost of path enumeration grows with the number of high-degree entities admitted to the traversal. Hub suppression thus trades connections that run through high-degree entities for tractability. Beyond these limitations, we see the reinterpretation of MHQA datasets as a more general contribution. A benchmark that provides question decompositions and intermediate answers can be reread as a relation inference benchmark. With this, \method opens several directions for future work, including extending the setting to branching questions with multiple anchors, purpose-built relation inference benchmarks, and retrievers trained directly on chain rather than answer supervision.

\newpage
\subsection*{AI use statement}
In this work, we used generative AI tools for tasks requiring mandatory disclosure. As outlined in section~\ref{sec:benchmark_construction}, we used LLMs to reformat the dataset MuSiQue by extracting anchor entities from decomposition questions. Our methodology was developed iteratively, including LLM feedback. To implement \method and baseline models, we used the support of LLMs for writing code. For manuscript editing, we used generative AI to improve clarity and readability, to help identify potentially relevant literature and assess the coverage of related work. All AI-assisted outputs were subsequently reviewed by the authors: Suggested references were manually verified against the original sources, manuscript changes were inspected and approved by the authors, and AI-assisted code was reviewed for correctness. The authors take full responsibility for the final content of this work, including all text, claims, results, and artifacts produced with the assistance of generative AI.

\subsection*{Reproducibility statement}

To support reproducibility, we document every component needed to rerun our experiments. The full \method pipeline is specified in Section~\ref{sec:method}, with retrieval summarised in Algorithm~\ref{alg:conrag_retrieval} and all hyperparameters fixed across datasets and listed in Table~\ref{tab:params}. Our benchmarks are derived from the public validation splits of MuSiQue and 2WikiMultiHopQA, and Section~\ref{sec:benchmark_construction} together with Appendix~\ref{app:datasets} describes the question selection, anchor and bridge extraction, and corpus pooling in detail. Baselines are taken from their public repositories and configured as described in Appendix~\ref{app:baselines_imp}, and every prompt used in the pipeline, from benchmark construction to evaluation, is included in Appendix~\ref{app:llm_prompts}. Computational requirements are reported in terms of token cost for indexing and generation (Table~\ref{tab:prompt_comparison_sizes}). Code and derived benchmark data are available in an anonymous repository.


\subsubsection*{Acknowledgments}

 SL is supported by the Swiss State Secretariat for Education, Research, and Innovation (SERI) under contract number MB22.00047.

\bibliographystyle{iclr2026_conference}
\bibliography{custom}

\appendix
\clearpage

\section{Extended Related Work}
\label{app:ext_rel_work}

\paragraph{Graph-based RAG variants.}
Beyond GraphRAG, LightRAG, and HippoRAG, a number of methods adapt the graph or the retrieval step. HyperGraphRAG~\citep{luo2025hypergraphrag} represents $n$-ary relations as hyperedges, and RAPTOR~\citep{sarthi2024raptor} replaces the graph with a tree of recursive summaries. KG$^2$RAG~\citep{zhu2025knowledge} expands initially retrieved chunks with graph neighbours. KET-RAG~\citep{huang2025ket} pairs a small LLM-extracted skeleton with a keyword-document bipartite graph to reduce indexing cost. All of these still rely on an LLM pass over the corpus for entity and relation extraction or summarisation.

\paragraph{Cost-efficient graph construction for RAG.}
Several more works reduce the LLM cost of building graphs for retrieval. CG-RAG~\citep{hu2025cg} avoids extraction entirely by navigating citation links between academic papers. \citet{min2025towards} replace the LLM with SpaCy's dependency parser for graph construction, and \citet{wang2026pruning} use REBEL as a triplet extractor to build multiple smaller triplet graphs that are orchestrated during retrieval. Without any graph, Multi-Meta-RAG~\citep{poliakov2024multi} uses extracted metadata to prefilter documents over several steps. \method requires no external structure such as citation links, and no relation extractor at all: The graph is relation-free, its
entity nodes are surface mentions mined from cooccurrence statistics and qualified by an LLM. Its retrieval also
serves a different goal. \method searches for paths between two known endpoints, which is what makes the intermediate entities and documents available as output.

\section{Extended Experimental Setting}
\label{app:exp_setup}

\subsection{Additional \method Implementation Details}

\begin{table}[h!]
  \centering
  \small
  \caption{Free parameters of \method and the values used
    in all experiments.}
  \label{tab:params}
  \begin{tabular}{@{}llrp{0.45\linewidth}@{}}
    \toprule
    Stage & Symbol & Value & Role \\
    \midrule
    \multirow{2}{*}{Entity qualification}
      & $n_{\max}$          & 5       & maximum phrase length in
                                        candidate mining \\
      & $B$                 & 200
      & batch size for containment-aware
                                        adjudication \\
    \midrule
    Indexing
      & $\theta$            & $0.8$   & embedding similarity threshold for
                                        entity node merging \\
    \midrule
    \multirow{3}{*}{Traversal}
      & $k_{\mathrm{seed}}$ & 20       & number of initial vector retrieval seed documents per question \\
      & $k_{\mathrm{fallback}}$ & 4       & number of vector retrieval fallback documents for an entity not contained in a document \\      
      & $p_{\mathrm{hub}}$  & $0.9$   & degree quantile above which an
                                        entity is suppressed \\
      & $L_{\max}$          & $5$     & maximum path length in documents \\
    \midrule
    \multirow{2}{*}{Selection}
      & $k_{\mathrm{path}}$ & $5$     & paths retained per length \\
      & $\mathcal{L}$       & $\{3,4,5\}$ & target path lengths \\
    \bottomrule
  \end{tabular}
\end{table}

Table~\ref{tab:params} lists the free parameters of \method and the values used
in all experiments. Entity qualification is governed by the maximum phrase length
$n_{\max}$ and the filtering batch size $B$, indexing by the merging threshold
$\theta$, traversal by the seed counts $k_{\mathrm{seed}}$ and
$k_{\mathrm{fallback}}$, the hub quantile $p_{\mathrm{hub}}$ and the maximum path
length $L_{\max}$, and path selection by $k_{\mathrm{path}}$ and the target
lengths $\mathcal{L}$. These values are fixed across both benchmarks rather than
tuned per dataset. With $p_{\mathrm{hub}}=0.9$, entities whose document frequency
exceeds the 90th percentile of the degree distribution are excluded from path
enumeration. 

\subsection{Baselines Implementation}
\label{app:baselines_imp}

\subsubsection{Na\"{i}ve RAG}
\label{app:baselines_imp:naive}

Na\"{i}ve RAG~\citep{lewis2020retrieval} serves as the non-graph reference point: Single-shot dense retrieval over the same pooled corpus, without query rewriting, reranking or iterative retrieval. Every pooled document is indexed as one unit. documents are embedded with \textit{text-embedding-3-small} using $1{,}536$ dimensions. We store them in an exact FAISS inner-product index. At query time the relation question $q$ is embedded with the same encoder and the top $k=10$ documents by $\cos(\mathbf{v}_q,\mathbf{v}_d)$ are retrieved. The retrieved documents are inserted as a numbered list of title-text pairs into the shared generation prompt of Appendix~\ref{app:llm_prompts_generation}. The response is generated by \textit{GPT-4o}. Indexing requires no LLM call, which is why no construction cost is reported in Table~\ref{tab:prompt_comparison_sizes}.

\subsubsection{LightRAG} 
\label{app:baselines_imp:lightrag}

LightRAG~\citep{guo2024lightrag} serves as one of the graph-based reference points. Entity and relationship extraction is performed with \textit{GPT-4o-mini}. Entities, relationships and chunks are embedded with \textit{text-embedding-3-small} using $1{,}536$ dimensions and stored in LightRAG's native NanoVectorDB and NetworkX backends. Queries are run in hybrid mode, with retrieval capped at $k=10$ documents. The retrieved context is inserted into the shared generation prompt and the response is generated by \textit{GPT-4o}.

\subsubsection{PathRAG}
\label{app:baselines_imp:pathrag}

PathRAG~\citep{chen2026pathrag} is the baseline whose retrieval mechanism is closest to \method's own. Since PathRAG and LightRAG share their underlying graph~\citep{chen2026pathrag}, we use the same NanoVectorDB index and NetworkX graph for MuSiQue and 2WikiMultiHopQA which explains the equal construction cost in Table~\ref{tab:prompt_comparison_sizes}. Vector search retrieves the $T{=}10$ entity nodes and relationship edges most similar to these keywords, forming the candidate pool $V_q$. This pool is then pruned to the $K{=}15$ highest-scoring relational paths. The per-hop decay rate is set at $\alpha{=}0.8$
and the decision retention threshold is set at $0.3$. The response is generated by \textit{GPT-4o}.

\subsubsection{HippoRAG 2}
\label{app:baselines_imp:hipporag2}

HippoRAG 2~\citep{gutierrez2025rag} serves as the graph-based RAG reference point designed for multi-hop reasoning. \textit{GPT-4o-mini} performs entity and relation extraction. \textit{text-embedding-3-small} using $1{,}536$ dimensions embeds both documents and extracted facts. Synonymy edges are added between entity nodes whose embeddings exceed a cosine similarity threshold of 0.8. We retrieve the top $k=10$ documents via PPR with a damping factor of 0.5. We use \textit{GPT-4o} as recognition memory and as generator and use the shared generation prompt with the retrieved documents.

\subsection{Datasets}
\label{app:datasets}

\subsubsection{Original Datasets}
\label{app:datasets:original}

\paragraph{MuSiQue.}
MuSiQue~\citep{trivedi2022musique} is built from single-hop questions of five Wikipedia-based reading comprehension datasets. Two questions are composed if the answer of the first is a named entity mentioned in the second. A composition is kept only if no subquestion can be answered without its predecessor's answer. Each question is described by the graph of its subquestions, which has two to four hops and one of six shapes. In the three linear shapes every subquestion depends only on the answer of the one before it. The question names a single starting entity, and the intermediate answers form one ordered chain to the final answer: \emph{What currency is used where Billy Giles died?} resolves \emph{Billy Giles} $\rightarrow$ \emph{Belfast} $\rightarrow$ \emph{Northern Ireland} $\rightarrow$ \emph{pound
sterling}. In the three branching shapes a subquestion depends on the answers of two predecessors that are independent of each other. \emph{When was the first establishment that McDonaldization is named after opened in the country Horndean is located?} separately resolves \emph{McDonaldization} $\rightarrow$ \emph{McDonald's} and \emph{Horndean} $\rightarrow$ \emph{England}, and only their combination (\emph{When did the first McDonald's open in England?}) yields the answer \emph{1974}. Such a question starts from two entities, and its intermediate answers are not connected to each other, so they form two converging paths rather than one chain between an anchor and a target. Every question ships with its decomposition containing subquestions, intermediate answers, one supporting paragraph per hop and a context of 20 paragraphs including hard distractors. The validation split answerable questions has 2,417 questions containing 1,252 two-hop, 760 three-hop and 405 four-hop questions.

\paragraph{2WikiMultiHopQA.}
2WikiMultiHopQA~\citep{ho2020constructing} is generated from templates over Wikidata triples that are aligned with the summary paragraphs of the corresponding Wikipedia articles. It has four question types: \textit{comparison}, \textit{inference}, \textit{compositional} and \textit{bridge comparison}. Compositional and inference questions are both built from two chained triples
$(e, r_1, e_1)$ and $(e_1, r_2, e_2)$, where the paragraph of $e$ must mention
the bridge $e_1$ and the paragraph of $e_1$ must mention the answer $e_2$.
Inference questions additionally collapse $r_1$ and $r_2$ into a single relation
through one of 28 logical rules (e.g.\ \emph{father} $\wedge$ \emph{father}
$\Rightarrow$ \emph{paternal grandfather}), so the question does not hint at the
bridge. Each question is annotated with its evidence triples and a context of 10
paragraphs, the gold paragraphs plus tf-idf retrieved distractors. The dataset
has 192,606 questions, where 12,576 questions belong to the validation set.

Both benchmarks thus annotate the intermediate entities of every question
explicitly, which is what allows them to be reread as relation inference
benchmarks in Section~\ref{app:datasets:construction}.
 
\subsubsection{Processing for multi-hop relation inference}
\label{app:datasets:construction}
 
\paragraph{Scope.}
Of the 2,417 MuSiQue validation questions we keep the linear compositions \textit{2hop},
\textit{3hop1} and \textit{4hop1} and drop the branching types \textit{3hop2}, \textit{4hop2}
and \textit{4hop3}, which start from two independent anchors. This results in 2066 questions. Of the 12,576 2WikiMultiHopQA
questions we keep \textit{compositional} and \textit{inference} resulting in 6785 questions. \textit{comparison} and \textit{bridge comparison} compare two anchors and have no bridge entity.

\vspace{-0.4cm}
\paragraph{Anchor, target, bridges and chain.}
The target $b$ is the answer of the source question, the anchor $a$ the entity its reasoning chain starts
from. In 2WikiMultiHopQA every question carries a list of (subject, relation, object) evidence triples:
The subject of the first triple is the anchor, the objects of all but the last triple are the bridges,
and the triples in order give the gold links. In MuSiQue the decomposition lists one subquestion per hop.
If the first subquestion has the structured form \texttt{<entity> >> <relation>} the anchor is parsed
directly. If the decomposition occurs to be a subquestion it is extracted by \textit{GPT-4o-mini} by using the prompt in Appendix~\ref{app:llm_prompts_extraction}. Intermediate answers are the bridges. Anchor, intermediate answers and target chained in decomposition order form the gold links. Each anchor-target pair is inserted into one of six paraphrases of \emph{How are \{anchor\} and \{target\} related?}, drawn uniformly at random.

\vspace{-0.4cm}
\paragraph{Anchor extraction quality.}
LLM extraction is the only non-deterministic step. The extracted anchor is a verbatim substring of the source question for 92.5\% of the affected questions and occurs verbatim in the supporting document of the first decomposition step for 77.4\%. In a manual check of 100 random extractions, 92 named the entity the gold chain starts from.
 
\begin{casebox}{Worked example: 2WikiMultiHopQA [compositional]}
\textbf{Source:} \emph{Where was the director of film Whispers (1990 Film) born?}\\
\textbf{Evidence:} \\
{}[Whispers - director - Douglas Jackson] \\
{}[Douglas Jackson - place of birth - Montreal, Quebec]

\tcbline

\textbf{Anchor} \emph{Whispers}, \textbf{Target} \emph{Montreal, Quebec}, \textbf{Bridge} \{Douglas Jackson\}.\\
\textbf{Links:} \\
{}[\texttt{Whispers -> Douglas Jackson}] \\
{}[\texttt{Douglas Jackson -> Montreal, Quebec}] \\
\textbf{Relation question:} \emph{How are Whispers and Montreal, Quebec related?}
\end{casebox}

\begin{casebox}{Worked example: MuSiQue [4-hop]}
\textbf{Source:} \emph{The German priest, who wanted Jean Hengen's religious denomination to reform, preached a sermon on Marian devotion a month before his death in what German state?}\\
\textbf{Decomposition:} \\
(1) \texttt{Jean Hengen >> religion} $\Rightarrow$ Catholic Church \\
(2) \emph{who wanted \#1 to reform} $\Rightarrow$ Martin Luther \\
(3) \emph{Where did \#2 preach a sermon on Marian devotion a month before his death?} $\Rightarrow$ Wittenberg \\
(4) \texttt{\#3 >> located in the administrative territorial entity} $\Rightarrow$ Saxony-Anhalt.

\tcbline

\textbf{Anchor} \emph{Jean Hengen}, \textbf{Target} \emph{Saxony-Anhalt}, \textbf{Bridges} \{Catholic Church, Martin Luther, Wittenberg\}.\\
\textbf{Links:} \\
{}[\texttt{Jean Hengen -> Catholic Church}] \\
{}[\texttt{Catholic Church -> Martin Luther}] \\
{}[\texttt{Martin Luther -> Wittenberg}] \\
{}[\texttt{Wittenberg -> Saxony-Anhalt}] \\
\textbf{Relation question:} \emph{What links Jean Hengen to Saxony-Anhalt?}
\end{casebox}
Chain recall and precision are computed against these link strings after lower-casing and accent
folding, bridge contains match against the bridge strings.
 
\subsubsection{Question statistics}
\label{app:datasets:questions}
 
An $n$-hop source question yields $n$ gold links, $n{-}1$ bridges and $n$ gold documents. MuSiQue
contains 1,252 with two, 568 with three and 246 questions with four hops. All
6,785 2WikiMultiHopQA questions are 2-hop with a single bridge. For each dataset three samples of 1,000 questions are drawn with seeds 123, 456, and 789.
 
\subsubsection{Pooled corpora}
 
The retrieval corpus of each dataset pools the context paragraphs of all validation questions and
deduplicates on whitespace collapsed title and text. A MuSiQue question originally comes with 20 context paragraphs of which two to four are supporting. A 2WikiMultiHopQA question with 10 of which two are supporting. After pooling every question is answered against the full corpus rather than its own paragraph bundle. The full corpus for MuSiQue contains approximately 26.3k documents and 56.7k for 2WikiMultiHopQA.
 
\subsubsection{Limitations of the derived benchmark}
\label{app:datasets:limitations}

\begin{table}[h!]
\centering
\small
\caption{\method on full vs clean question subsets. Values are mean $\spm$ SD across seeds.}
\begin{tabular}{llcccc}
\hline
Dataset & Subset & Avg.\ $|\mathcal{Q}|$ & BCM & CR & CP \\
\hline
MuSiQue & full  & 1000.0 & 71.7 $\spm$ 1.3 & 28.4 $\spm$ 1.3 & 34.5 $\spm$ 2.4 \\
MuSiQue & clean & 840.3  & 74.8 $\spm$ 0.3 & 32.4 $\spm$ 0.7 & 39.2 $\spm$ 0.4 \\
\hline
2WikiMultiHopQA & full  & 1000.0 & 81.0 $\spm$ 1.4 & 49.4 $\spm$ 1.3 & 48.1 $\spm$ 0.9 \\
2WikiMultiHopQA & clean & 985.3  & 81.1 $\spm$ 1.3 & 49.5 $\spm$ 1.3 & 48.2 $\spm$ 1.2 \\
\hline
\end{tabular}

\label{app:clean_questions}
\end{table}

Gold chains are only as complete as the source decompositions. A valid alternative path between anchor and target is scored as a false link. Moreover, not every gold link is supported by the document text. We
call an entity grounded in a document if all of its tokens occur in that document after lower-casing and accent folding, which is the most tolerant matching rule the retriever itself applies. On MuSiQue the
anchor is not grounded in any gold document for 163 of the 2,066 questions. The target is not grounded for 6 questions. A bridge is not grounded in both documents it connects for 189 of the 3,126 MuSiQue bridges affecting 186 questions and for 95 of the 6,785 2WikiMultiHopQA bridges. Taken together, 312 MuSiQue questions and 95 2WikiMultiHopQA questions contain at least one gold link that no document supports. As shown in Table \ref{app:clean_questions} restricting the evaluation to the 1,754 grounded MuSiQue questions raises every metric for \method.

\section{Comprehensive Multi-hop Relation Inference Performance}
\label{app:add_results}
\subsection{Detailed Results Overview}

\begin{table*}[h]
\centering
\small
\setlength{\tabcolsep}{5pt}
\caption{Hop-stratified relation retrieval performance across systems. BCM = Bridge Contains Match Accuracy, CR = chain recall, and CP = chain precision. Values are mean $\spm$ standard deviation across available seeds. Bold indicates the best score within each dataset-hop block and metric.}
\begin{tabular}{lll l ccc}
\hline
\textit{Dataset} & \textit{Hops} & \textit{Model} & \textit{BCM} & \textit{CR} & \textit{CP} \\
\hline
2WikiMultiHopQA & 2 & \method & \textbf{81.0\% $\spm$ 1.4} & \textbf{49.4\% $\spm$ 1.2} & \textbf{48.1\% $\spm$ 0.9} \\
 &  & HippoRAG 2 & 67.0\% $\spm$ 1.0 & 40.2\% $\spm$ 0.9 & 34.9\% $\spm$ 2.1 \\
 &  & LightRAG & 72.1\% $\spm$ 0.2 & 40.2\% $\spm$ 1.0 & 35.7\% $\spm$ 1.0 \\
 &  & PathRAG & 64.9\% $\spm$ 0.2 & 30.4\% $\spm$ 1.5 & 35.8\% $\spm$ 1.8 \\
 &  & Na\"{i}ve RAG & 61.6\% $\spm$ 1.5 & 33.4\% $\spm$ 2.4 & 40.1\% $\spm$ 2.3 \\
\hline
MuSiQue & 2 & \method & \textbf{84.2\% $\spm$ 1.2} & \textbf{43.7\% $\spm$ 0.9} & 38.5\% $\spm$ 1.5 \\
 &  & HippoRAG 2 & 72.3\% $\spm$ 2.1 & 35.2\% $\spm$ 0.7 & \textbf{40.2\% $\spm$ 2.0} \\
 &  & LightRAG & 72.1\% $\spm$ 2.6 & 30.3\% $\spm$ 0.3 & 31.1\% $\spm$ 0.5 \\
 &  & PathRAG & 71.9\% $\spm$ 1.6 & 26.7\% $\spm$ 0.3 & 36.9\% $\spm$ 1.4 \\
 &  & Na\"{i}ve RAG & 72.2\% $\spm$ 0.8 & 29.8\% $\spm$ 0.6 & 30.9\% $\spm$ 0.8 \\
\hline
 & 3 & \method & \textbf{57.6\% $\spm$ 0.3} & \textbf{9.4\% $\spm$ 0.2} & \textbf{25.5\% $\spm$ 0.5} \\
 &  & HippoRAG 2 & 43.8\% $\spm$ 1.5 & 5.9\% $\spm$ 0.6 & 24.9\% $\spm$ 1.7 \\
 &  & LightRAG & 40.1\% $\spm$ 2.7 & 5.2\% $\spm$ 0.9 & 19.0\% $\spm$ 2.1 \\
 &  & PathRAG & 40.8\% $\spm$ 1.0 & 3.0\% $\spm$ 0.2 & 19.4\% $\spm$ 2.0 \\
 &  & Na\"{i}ve RAG & 42.8\% $\spm$ 0.1 & 4.3\% $\spm$ 0.7 & 16.6\% $\spm$ 3.1 \\
\hline
 & 4 & \method & \textbf{36.8\% $\spm$ 2.6} & \textbf{3.8\% $\spm$ 0.5} & 12.0\% $\spm$ 0.8 \\
 &  & HippoRAG 2 & 31.8\% $\spm$ 4.1 & 3.2\% $\spm$ 0.7 & \textbf{17.1\% $\spm$ 1.8} \\
 &  & LightRAG & 28.4\% $\spm$ 2.7 & 3.7\% $\spm$ 0.8 & 16.5\% $\spm$ 2.7 \\
 &  & PathRAG & 30.6\% $\spm$ 2.7 & 2.0\% $\spm$ 1.0 & 15.5\% $\spm$ 4.5 \\
 &  & Na\"{i}ve RAG & 31.9\% $\spm$ 2.6 & 3.1\% $\spm$ 0.5 & 12.9\% $\spm$ 1.6 \\
\hline
\end{tabular}
\label{tab:hop_breakdown_br_cr_cp}
\end{table*}

Table~\ref{tab:hop_breakdown_br_cr_cp} stratifies the results of
Table~\ref{tab:bridge-chain} by hop count. \method leads BCM in every block, by $11.9$, $13.8$ and $4.9$ points on two-, three- and four-hop MuSiQue questions, with a relative decay comparable to the baselines. The chain metrics separate the systems clearly on two-hop questions, where \method gains points in chain precision and chain recall on 2WikiMultiHopQA and outperforms HippoRAG 2 on MuSiQue in chain recall. At three and four hops every system recovers fewer than one in ten and one in twenty gold links. The the nominal lead of LightRAG at two hops on precision, lies within one standard deviation. At four hops \method's chain precision lies below baselines. Notably, \method still names most of the gold bridges on four-hop questions and leads in chain recall.\looseness-1

\subsection{Win Case Analysis}

To illustrate how \method's path enumeration translates into the gains reported above, we contrast its output with that of HippoRAG\,2, the strongest baseline on chain recall, on a two-hop MuSiQue question. The example is representative of a win pattern in our runs. Both systems retrieve the anchor document, but only \method reaches the bridge document, since the shared entity \emph{Palembang} links the anchor document to the target document in the entity-document graph and the resulting path is scored highly by query support. HippoRAG\,2 instead spreads activation from the query entities toward semantically related wartime documents that mention neither endpoint jointly. As the bridge document is the only document that ties the birthplace of the anchor to the date of the target, its presence in the context decides whether a chain can be emitted at all.

\begin{casebox}{Win case: What links Marzuki Alie to 15 February 1942?}
\textbf{Gold chain:} Marzuki Alie $\rightarrow$ Palembang $\rightarrow$ 15 February 1942. \\
\textbf{Gold documents:} \textbf{\emph{Battle of Palembang}}, \textbf{\emph{Marzuki Alie}}. \\
\textbf{Bridge entity:} Palembang, described in \textbf{\emph{Battle of Palembang}}. Gold documents are set in bold wherever they appear below.

\tcbline

\textbf{\method.} Retrieved documents: \textbf{\emph{Battle of Palembang}}, \textbf{\emph{Marzuki Alie}}, \emph{Tengku Tengah Zaharah Mosque}, \emph{Choo Hoey}, \emph{Arie Supriyatna}, \emph{1979 Garuda Fokker F28 crash}, \emph{Mohd Marzuki Yusof}, \emph{Sang Nila Utama}, \emph{Denny Januar Ali}, \emph{Muhammad Shobran}, \emph{Southeast Asia}, \emph{Pacific War}. \\
\textbf{Predicted chain:} Marzuki Alie $\rightarrow$ Palembang $\rightarrow$ 15 February 1942.
\begin{quote}\itshape
Marzuki Alie was born in Palembang, South Sumatra, on 6 November 1955. Palembang is also the location of the Battle of Palembang, which occurred during the Pacific theatre of World War II from 13--15 February 1942. The connection between Marzuki Alie and 15 February 1942 is through the shared location of Palembang, which ties Marzuki Alie's birthplace to the historical event that took place there on that date.
\end{quote}

\tcbline

\textbf{HippoRAG~2.} Retrieved documents (top ten, the documents its prompt carries): \textbf{\emph{Marzuki Alie}}, \emph{Julius Ellsberry}, \emph{Battle of the Tenaru}, \emph{Abdul Ali Mazari}, \emph{Military history of the United States during World War II}, \emph{September 11 attacks}, \emph{Abu Yahya al-Libi}, \emph{Larry Alcala}, \emph{United Defense M42}, \emph{Marshall Islands}. \\
\textbf{Predicted chain:} none returned.
\begin{quote}\itshape
The provided documents do not establish any meaningful connection between Marzuki Alie and the date 15 February 1942. Marzuki Alie is described as a political figure from Indonesia, serving as the speaker of the People's Representative Council and a member of the Partai Demokrat. The date 15 February 1942 is not mentioned in the context, nor is there any event or entity linked to Marzuki Alie that connects him to this specific date. The documents retrieved do not support any relationship beyond coincidental term overlap.
\end{quote}
\end{casebox}

\noindent The bridge document \emph{Battle of Palembang} reaches \method's context but not HippoRAG~2's, whose ten documents are drawn instead from topically adjacent wartime material. The failure is in retrieval rather than generation: The connecting fact is absent from the context, so no amount of reasoning over it recovers the chain.

\subsection{False Negative Analysis}

\begin{table}[h!]
  \centering
  \footnotesize
  \setlength{\tabcolsep}{4pt}
\caption{Construction of the false-negative pool (1000 questions per system). \textbf{Match} = at least one gold link reproduced exactly; \textbf{Abst.} = no chain emitted; \textbf{Deny} = a chain was emitted but the prose denies a connection; \textbf{Near} = a gold link recovered under a different surface form; \textbf{Unsp.} = the predicted chain does not span anchor to target. \textbf{Alt} is the remaining pool of alternative-path candidates.}
  \begin{tabular}{llrrrrrr}
    \toprule
    Dataset & System & Match & Abst. & Deny & Near & Unsp. & Alt \\
    \midrule
    \multirow{5}{*}{\textit{MuSiQue}} & \method & 380 & 403 & 25 & 52 & 13 & 127 \\
     & HippoRAG 2 & 300 & 511 & 13 & 59 & 10 & 107 \\
     & LightRAG & 276 & 548 & 18 & 36 & 26 & 96 \\
     & Na\"{i}ve RAG & 253 & 549 & 26 & 65 & 2 & 105 \\
     & PathRAG & 238 & 621 & 11 & 30 & 25 & 75 \\
    \midrule
    \multirow{5}{*}{\textit{2WikiMultiHopQA}} & \method & 643 & 173 & 13 & 60 & 4 & 107 \\
     & HippoRAG 2 & 489 & 301 & 37 & 52 & 19 & 102 \\
     & LightRAG & 496 & 241 & 17 & 96 & 14 & 136 \\
     & Na\"{i}ve RAG & 369 & 417 & 36 & 49 & 12 & 117 \\
     & PathRAG & 353 & 397 & 20 & 78 & 13 & 139 \\
    \bottomrule
  \end{tabular}
  \label{tab:fn_pool_mixed}
\end{table}

\begin{table}[h!]
  \centering
  \footnotesize
  \setlength{\tabcolsep}{4pt}
\caption{Classification of the alternative-path candidates and blind validation. \textbf{Alias}: the chain reaches the gold bridge under a different surface form. \textbf{Short}: the chain bypasses the gold bridge and links the endpoints directly. \textbf{Degen}: an intermediate node is a description or a restatement of the anchor or target rather than a distinct entity. \textbf{Distinct} cases are judged blind against that system's own retrieved documents; a case is \textbf{Valid} only if every link is supported, the chain connects anchor to target, and every step is substantive.}
  \begin{tabular}{llrrrrrrr}
    \toprule
    & & & \multicolumn{3}{c}{Excluded} & & & \\
    \cmidrule(lr){4-6}
    Dataset & System & Alt & Alias & Short & Degen & Distinct & Valid & Rate \\
    \midrule
    \multirow{5}{*}{\textit{MuSiQue}} & \method & 127 & 24 & 5 & 18 & 80 & 70 & 87.5\% \\
     & HippoRAG 2 & 107 & 20 & 8 & 16 & 63 & 58 & 92.1\% \\
     & LightRAG & 96 & 18 & 8 & 11 & 59 & 40 & 67.8\% \\
     & Na\"{i}ve RAG & 105 & 24 & 6 & 11 & 64 & 55 & 85.9\% \\
     & PathRAG & 75 & 16 & 7 & 11 & 41 & 20 & 48.8\% \\
    \midrule
    \multirow{5}{*}{\textit{2WikiMultiHopQA}} & \method & 107 & 19 & 10 & 17 & 61 & 58 & 95.1\% \\
     & HippoRAG 2 & 102 & 15 & 4 & 18 & 65 & 61 & 93.8\% \\
     & LightRAG & 136 & 45 & 8 & 15 & 68 & 61 & 89.7\% \\
     & Na\"{i}ve RAG & 117 & 29 & 12 & 19 & 57 & 50 & 87.7\% \\
     & PathRAG & 139 & 44 & 6 & 16 & 73 & 48 & 65.8\% \\
    \bottomrule
  \end{tabular}
  \label{tab:fn_classify_mixed}
\end{table}

Since the evaluation protocol accepts only the gold decomposition as
reference, a chain that connects anchor and target through a different but
valid route is scored as an error. To estimate how much of the reported
error is of this kind, we analyse the questions of the first seed from the MuSiQue and 2WikiMultiHopQA relation samples on which a system reproduces no gold link. Table~\ref{tab:fn_pool_mixed} partitions these questions into abstentions, denials of a connection, near matches under a different surface form, chains that do not span both endpoints and the remaining pool of alternative-path candidates. Table~\ref{tab:fn_classify_mixed} further removes candidates that reach the
gold bridge under an alias, bypass it or route through a degenerate node,
and validates the distinct remainder blind against the retrieved documents of
the respective system. Two findings follow. First, abstentions dominate the false negatives of every system, so the primary failure mode is missing evidence rather than a wrong chain. Second, valid alternative chains are common but do not change the ranking. For \method, $70$ of $80$ distinct candidates on MuSiQue and $58$ of $61$ on 2WikiMultiHopQA are supported by
its own context. The baselines gain similar absolute amounts, while LightRAG and PathRAG show markedly lower validity rates, indicating that their deviating chains are more often unsupported by the retrieved documents than genuinely alternative. The two cases below illustrate valid alternatives that the metric rejects.

\begin{casebox}{Case 1: What links Australia to Lyons' death in 1939?}
\textbf{Gold path:} Australia $\rightarrow$ Robert Menzies $\rightarrow$ Lyons' death in 1939. \\
\textbf{Alternative path:} Australia $\rightarrow$ Joseph Lyons $\rightarrow$ Lyons' death in 1939.

\tcbline

\textbf{\method} explains the connection as follows:
\begin{quote}\itshape
Joseph Lyons was a prominent political figure in Australia, serving as the leader of the United Australia Party (UAP) and as Prime Minister. His death in 1939 marked a significant event in Australian political history, as it led to Robert Menzies assuming the Prime Ministership. Thus, the connection between Australia and Lyons' death in 1939 is through his role as a key political leader in the country.
\end{quote}
\end{casebox}

\noindent The gold chain routes through Robert Menzies, who succeeded Lyons. The alternative routes through Joseph Lyons himself, the person whose death the question names. It is the more direct reading, and the metric scores it zero because neither link string matches.

\begin{casebox}{Case 2: Explain the relationship between Mickey Mouse and Walt Disney.}
\textbf{Gold path:} Mickey Mouse $\rightarrow$ Minnie Mouse $\rightarrow$ Walt Disney. \\
\textbf{Alternative path:} Mickey Mouse $\rightarrow$ Ub Iwerks $\rightarrow$ Walt Disney.

\tcbline

\textbf{\method} explains the connection as follows:
\begin{quote}\itshape
Collaboration with Ub Iwerks: Ub Iwerks, a close collaborator of Walt Disney, was instrumental in designing Mickey Mouse. Iwerks created the initial sketches of the character, inspired by earlier ideas and sketches of mice around a photograph of Walt Disney. This collaboration ties Mickey Mouse to Walt Disney through their shared creative process.
\end{quote}
\end{casebox}

\noindent The gold chain routes through Minnie Mouse, a second fictional character. The alternative routes through Ub Iwerks, who drew the original sketches of Mickey Mouse and co-created the character with Disney.

\section{Mining and Retrieval Studies}
\label{app:ablations}

\subsection{Entity filtering}
\label{app:llm_filtering}
  \begin{figure}[h!]
    \centering
    \includegraphics[width=\linewidth]{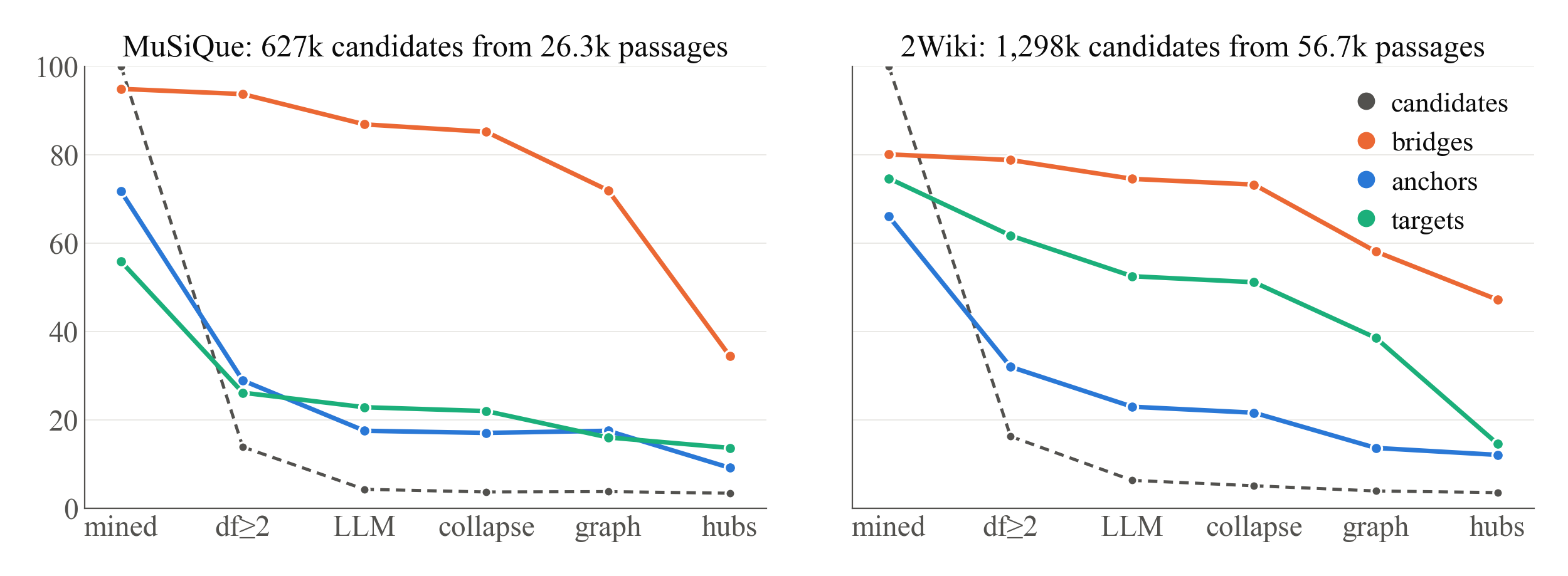}
    \caption{\textbf{Entity vocabulary funnel.} Share (\%) of mined candidates
    (dashed) and of gold bridges, targets and anchors that retain an entity node
    (solid) after each indexing stage on MuSiQue (left) and 2WikiMultiHopQA
    (right). The \emph{hubs} stage applies only to the traversal adjacency.}
    \label{fig:entity_funnel}
  \end{figure}

Figure~\ref{fig:entity_funnel} traces the entity vocabulary through
indexing and reports, after each step, the share of surviving candidates
together with the share of gold anchors, targets and bridges that still
have a node. The document frequency cut already removes about $85$\,\% of
the mined candidates. LLM filtering removes roughly two thirds of the
remainder, leaving about $4$\,\% and $6$\,\% of the initial vocabulary on
MuSiQue and 2WikiMultiHopQA. Bridge coverage, which is what path
enumeration depends on, is largely unaffected by these two steps. Anchors and targets lose more, but they are recovered by the endpoint constraint during seeding and need not be present in the graph.

\section{Overview of LLM Prompts}
\label{app:llm_prompts}
\subsection{Dataset Anchor Entity Extraction}
\label{app:llm_prompts_extraction}

\newtcblisting{promptverbatim}{%
  listing only, breakable,
  colback=gray!4, colframe=black!70,
  boxrule=0.6pt, arc=2pt,
  left=6pt, right=6pt, top=6pt, bottom=6pt,
  listing options={
    basicstyle=\ttfamily\scriptsize,
    breaklines=true, breakatwhitespace=true, breakindent=0pt, breakautoindent=false, upquote=true,
    columns=fullflexible, keepspaces=true,
  },
}
 
\begin{promptbox}
\begin{Verbatim}
You extract the one NAMED ENTITY that a multi-hop question starts from.
 
Multi-hop questions nest descriptions inside each other. Somewhere at the bottom of that nesting sits a single proper name - a person, place, work, organisation or product. That name is the anchor. Everything wrapped around it is description, not the answer.
 
"Who is the spouse of the Green performer?" -> Green performer
"Who gives out the prize named after the author of The Wealth of Nations?" -> The Wealth of Nations
"When was the death penalty abolished in the country near the country where the writer of The Book Thief is a citizen of?" -> The Book Thief
"Who is the creator of Mickey Mouse's spouse?" -> Mickey Mouse
"What county shares a border with the county where She Did It's performer was born?" -> She Did It
 
CRITICAL: return a NAME, never a description.
Never return anything containing "of the", "where", "which", "that", "named after", "the country", "the person", "the city". Those are descriptions of an entity, not the entity. If your answer needs a relative clause to make sense, you have the wrong span - go deeper into the nesting until you reach the bare proper name.
 
Rules:
1. Copy the name exactly as it appears, with all of its words and any leading "The" that is part of the title: "The Wealth of Nations", not "Wealth".
2. Almost always 1-4 words. More than that is a sign you returned a description.
3. Never return the thing being asked for, a property ("spouse", "founder"), or a bare category ("the company", "the film").
4. If several names appear, return the one the chain begins at - the most deeply nested.
 
Return ONLY a JSON array, one object per question, in order: [{"index": 0, "anchor": "..."}]
\end{Verbatim}
\end{promptbox}

This prompt is used once, during benchmark construction
(Section~\ref{sec:benchmark_construction}), and is not part of \method itself.
It applies only to MuSiQue questions whose first decomposition step is a
natural-language subquestion rather than the structured form
\texttt{<entity> >> <relation>}, from which the anchor can be parsed directly.
\textit{GPT-4o-mini} receives a batch of such subquestions and returns the single proper name the reasoning chain starts from for each. The few-shot examples and the rules steer the model towards the nested name and away from descriptive spans. The quality of the extracted anchors is assessed in Appendix~\ref{app:datasets:construction}.

\subsection{Entity Filtering}
\label{app:llm_prompts_filtering}
\begin{promptbox}
\begin{Verbatim}
You decide which words or short phrases are REAL NAMED ENTITIES.

A true entity should be a concrete value you could assign to an object slot, for example:
- person: "Emmanuel Macron"
- place: "Paris"
- organisation: "UNESCO"
- event/work/product: "World War II", "The Beatles"
- date/time value: "11 July", "January 19, 2018"
- language name as an object value: "French language", "Swahili"

Mark false if the item is generic, descriptive, grammatical, or not a specific referent.

Examples:
- true: "Paris", "Mozart", "Boeing", "United States", "World War II", "January 19, 2018"
- false: "main", "best", "series", "television", "river", "company", "building",
         "quickly", "running", "theory", "board of directors", "west of the river"

Input items (JSON array with index/text/capitalization_pct):
__ITEMS_JSON__

Return ONLY a JSON array with one object per input item, in order:
[{"index": 0, "entity": true}, {"index": 1, "entity": false}]
\end{Verbatim}
\end{promptbox}

This prompt implements the LLM-based entity filtering step of
Section~\ref{sec:method:entities} and accounts for the entire LLM token cost
of \method's indexing reported in Table~\ref{tab:prompt_comparison_sizes}.
\textit{GPT-4o} receives a containment-aware batch of up to $B$ mined
candidates as a JSON array that replaces the placeholder. Each item carries its index, its surface string
and its capitalisation rate across the corpus, which serves as a signal for distinguishing proper names from common words. The model returns one binary label per candidate. Only candidates labelled with \textit{true} enter the entity vocabulary.

\subsection{Generation}
\label{app:llm_prompts_generation}
\begin{promptbox}
\begin{Verbatim}
You describe how two entities are connected, using passages retrieved from a document graph.
You are given a set of PASSAGES (unique seed and path-selected documents) in this system prompt, and a QUESTION naming two entities in the user message. Use only the provided passages as evidence.
Your job is to give the reader the fullest possible account of how these two entities relate, **as far as the passages support it**. Not a verdict - an explanation.
Relationships can be:
- Example 1
  - Relationship question: "What is the connection between Mickey's Safari in Letterland and Walt Disney?"
  - Output-style chain:
    - `"Mickey's Safari in Letterland -> Mickey Mouse"`
    - `"Mickey Mouse -> Walt Disney"`
  - Explanation example:
    - "`Mickey's Safari in Letterland` is part of the `Mickey Mouse` series. `Mickey Mouse` was developed by `Walt Disney`.
- Example 2
  - Relationship question: "What links Zubly Cemetery to Richland County?"
  - Output-style chain:
    - `"Zubly Cemetery -> South Carolina"`
    - `"South Carolina -> Columbia"`
    - `"Columbia -> Forest Acres"`
    - `"Forest Acres -> Richland County"`
  - Explanation example:
    - "`Zubly Cemetery` is located in `South Carolina`, which has the capital `Columbia`. `Columbia` borders `Forest Acres`, which is located in `Richland County`."
- Example 3
  - Relationship question: "What links Houston Kid to Warner Music Group?"
  - Output-style chain:
    - `"Houston Kid -> Rodney Crowell"`
    - `"Rodney Crowell -> Warner Bros. Records"`
    - `"Warner Bros. Records -> Warner Music Group"`
  - Explanation example:
    - "`Houston Kid` is performed by `Rodney Crowell` whose record label is `Warner Bros. Records`. `Warner Bros. Records` belongs to `Warner Music Group`."
## Be comprehensive
Trace every thread the passages offer, not just the first or strongest:
All three are worth describing. Report each connection separately, strongest first, and say which kind it is so the reader can weigh it. A single entity pair often has several distinct threads running through different intermediates - find them all.
## Be diverse
Cover different *kinds* of connection, not several restatements of one. Prefer breadth across relation types over depth on one.
## Be empowering
Write so the reader can understand the connection and judge it themselves:
- Name the intermediates. "Connected through their producer" is weak; "connected through Steve Hillage, who performed on the album and is married to Miquette Giraudy" is useful.
- Give the specifics the passages contain - dates, roles, places, titles.
- Where the passages leave a gap, name the gap. Saying what is missing is part of a useful explanation.
## Stay grounded
Everything must come from the passages. Do not use outside knowledge, and do not invent a link to make a tidier story.
**A shared word is not itself a relationship.** Two passages both containing "1996" or "london" tells you they share a term, nothing more.
If the passages genuinely show nothing beyond coincidental term overlap, say so and describe what they do show - but look properly first.
In every case, only assert what the retrieved passages explicitly support.
Moreover, you shall explicitly chain related entities that relate to each other directly in the format outlined in the output
## Output
Return ONLY a JSON object:
{
  "chains": [
    [
      "Entity A -> Bridge 1",
      "Bridge 1 -> Bridge 2",
      "Bridge 2 -> Entity B"
    ]
  ],
  "explanation": "a comprehensive prose explanation of how the entities are related (or not), grounded only in the provided passages"
}
Each chain is a list of directed entity links in the form:
`"from_entity -> to_entity"`.
`chains` must contain one or more chains supported by the passages.
If no meaningful relationship is supported, return `"chains": []` and explain that only incidental overlap (or no overlap) was found.
Do not output any other keys besides `chains` and `explanation`.
\end{Verbatim}
\end{promptbox}

All systems, \method and the four baselines, share this generation prompt with
\textit{GPT-4o}. Only the passage block differs and holds the context
retrieved by the respective method, so differences in the results stem from
retrieval rather than from prompt design. The retrieved passages are placed in
the system prompt and the relation question in the user message. The model
returns a JSON object with two fields that mirror the output
$\mathcal{O}(q) = (\mathcal{C}, y)$ of Eq.~\ref{eq:output}. The grounding instructions allow the model
to return an empty chain list when the passages support no connection.

\subsection{Evaluation}
\label{app:llm_prompts_evaluation}

\begin{promptbox}
\begin{Verbatim}
---Role---
You are an expert tasked with evaluating two answers to the same question based on four criteria: \
**Comprehensiveness**, **Diversity**, and **Empowerment**.
---Goal---
You will evaluate two answers to the same question based on four criteria: **Comprehensiveness**, \
**Diversity**, and **Empowerment**.
- **Comprehensiveness**: How much detail does the answer provide to cover all aspects and details of the question?
- **Diversity**: How varied and rich is the answer in providing different perspectives and insights on the question?
- **Empowerment**: How well does the answer help the reader understand and make informed judgments about the topic?
For each criterion, choose the better answer (either Answer 1 or Answer 2) and explain why. Then, select an \
overall winner based on these three categories.
Here is the question: {question}
Here are the two answers:
**Answer 1:**
{answer1}
**Answer 2:**
{answer2}
Evaluate both answers using the three criteria listed above and provide detailed explanations for each criterion.
Output your evaluation in the following JSON format:
{{
    "Comprehensiveness": {{
        "Winner": "[Answer 1 or Answer 2]",
        "Explanation": "[Provide explanation here]"
    }},
    "Diversity": {{
        "Winner": "[Answer 1 or Answer 2]",
        "Explanation": "[Provide explanation here]"
    }},
    "Empowerment": {{
        "Winner": "[Answer 1 or Answer 2]",
        "Explanation": "[Provide explanation here]"
    }},
    "Overall Winner": {{
        "Winner": "[Answer 1 or Answer 2]",
        "Explanation": "[Summarize why this answer is the overall winner based on the three criteria]"
    }}
}}
\end{Verbatim}
\end{promptbox}
For the qualitative comparison in Table~\ref{tab:winrates} we adopt the pairwise judging prompt of LightRAG~\citep{guo2024lightrag}. For each relation question, \textit{GPT-4o-mini} is shown the explanation generated by \method and that of one baseline as Answer~1 and Answer~2. It selects a winner for comprehensiveness, diversity and empowerment, justifies each choice, and then names an overall winner based on the three criteria. Win rates are the share of questions on which \method is selected.

\end{document}